\documentclass[12pt]{article}

\usepackage[english]{babel}
\usepackage[a4paper,top=2cm,bottom=2cm,left=2cm,right=2cm,marginparwidth=1.75cm]{geometry}
\usepackage{caption}

\usepackage[utf8]{inputenc}
\usepackage[T1]{fontenc}
\usepackage{diagbox}
\usepackage{subfigure}
\usepackage{parskip}
\usepackage{float}
\usepackage{graphicx}
\usepackage{pythonhighlight}
\usepackage{enumerate}
\usepackage{amsfonts,amssymb}
\usepackage{amsmath}
\usepackage{graphicx}
\usepackage{indentfirst}
\usepackage[colorlinks=true, allcolors=blue]{hyperref}

\usepackage{afterpage}

\newcommand\myemptypage{
    \null
    \thispagestyle{empty}
    \addtocounter{page}{-1}
    \newpage
    }
\numberwithin{equation}{section}

\newcommand{\name}{Yang Bai}
\newcommand{\matrikel}{459272}
\newcommand{\thema}{\textbf{\Huge A Machine Learning accelerated geophysical fluid solver}}
\newcommand{\datum}{November 30, 2022} 

\begin{document}
\begin{figure}[h]
\begin{minipage}{0.2\textwidth}
 \centering
\includegraphics[width=\textwidth]{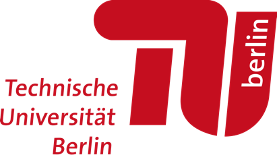}

\end{minipage}
\hfill
\begin{minipage}{0.3\textwidth}
 \centering
\includegraphics[width=\textwidth]{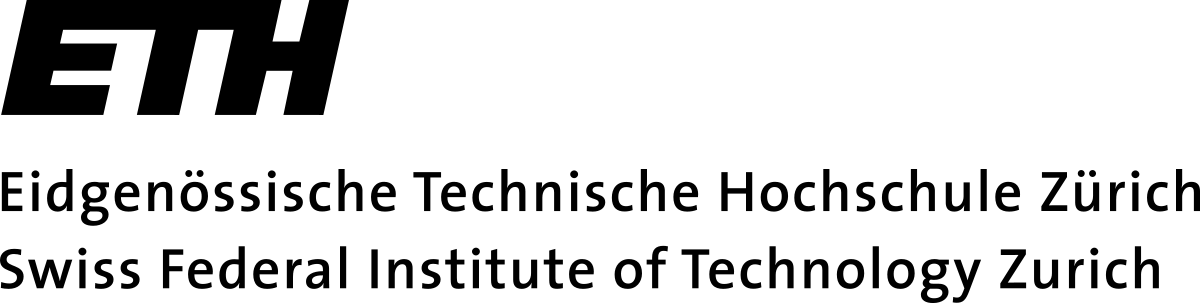}

\end{minipage}
\hfill
\begin{minipage}{0.3\textwidth}
 \centering
\includegraphics[width=\textwidth]{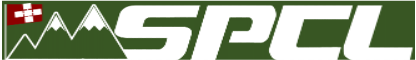}

\end{minipage}
\end{figure}

\vspace*{2cm}
\begin{center}
\LARGE
Master Thesis\\
\vspace{1.0cm}
\thema

\vspace{3.75cm}

\Large

\name\\
Matrikelnumber: \matrikel\\
\datum\\
\vspace{1.0cm}
\large
Completed in the Msc. Scientific Computing program at \\Institute for Mathematics of Technical University of Berlin\

%
\end{center}
\vspace{4cm}
Supervised by: \\
Prof. Dr. Torsten Höfler\\
\textit{Institute of High Performance Computing Systems, ETHz, Switzerland}\\~\\
Prof. Dr. Jörg Liesen\\
\textit{Institute of Mathematics, Technical University of Berlin, Germany}\\

\myemptypage
\begin{figure}[H]
\centering
\includegraphics[width=1\textwidth]{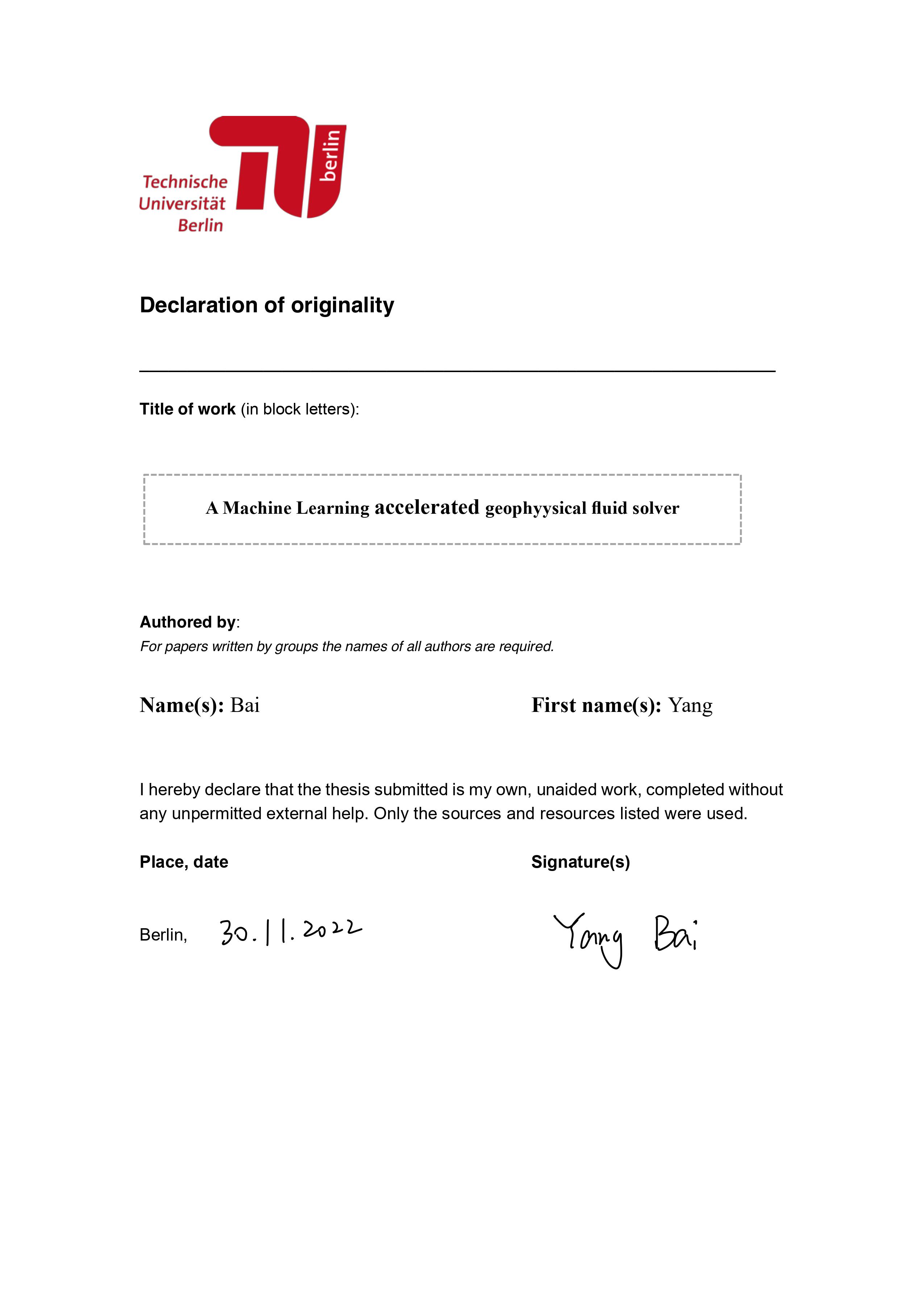}
\end{figure}

\newpage
\begin{center}
\section*{Abstract}
\end{center}
Machine learning methods have been successful in many areas, like image
classification and natural language processing. However, it still needs to be determined how to
apply ML to areas with mathematical constraints, like solving PDEs. Among
various approaches to applying ML techniques to solving PDEs, the data-driven discretization method presents a promising way of accelerating and
improving existing PDE solver on structured grids where it predicts the coefficients of quasi-linear stencils for computing values or derivatives of a function
at given positions. It can improve the accuracy and stability of low-resolution
simulation compared with using traditional finite difference or finite volume
schemes. Meanwhile, it can also benefit from traditional numerical schemes
like achieving conservation law by adapting finite volume type formulations.

In this thesis, we have implemented the shallow water equation and Euler equation classic solver under a different framework. Experiments show that 
our classic solver performs much better than the Pyclaw solver. Then we propose four different deep neural networks for the ML-based solver. The results indicate that two of these approaches could output satisfactory solutions.

\newpage
\begin{center}
\section*{Zusammenfassung}
\end{center}
Methoden des maschinellen Lernens haben sich in vielen Bereichen bewährt wie beispielsweise bei der Klassifizierung und der Verarbeitung natürlicher Sprache. Es muss jedoch noch geklärt werden, wie man ML auf Bereiche mit mathematischen Beschränkungen, wie der Lösung von PDEs, anwenden kann. Unter den verschiedenen Ansätzen zur Anwendung von ML-Techniken bei der Lösung von PDEs stellt die datengesteuerte Diskretisierungsmethode einen vielversprechenden Weg zur Beschleunigung und Verbesserung bestehender PDE-Solver auf strukturierten Gittern dar, bei denen sie die Koeffizienten quasi-linearer Schablonen zur Berechnung von Werten oder Ableitungen einer Funktion an bestimmten Positionen vorhersagt. So kann die Genauigkeit und Stabilität von Simulationen mit niedriger Auflösung im Vergleich zu herkömmlichen Finite-Differenzen- oder Finite-Volumen-Verfahren verbessert werden. Hierbei kann dies auch von traditionellen numerischen Schemata profitieren, wie z.B. der Umsetzung des Erhaltungssatzes durch die Anpassung von Formulierungen des Typs finites Volumen.

In dieser Arbeit wurde die Flachwassergleichung und der klassische Solver der Euler-Gleichung in einem anderen Rahmen implementiert. Experimente zeigen, dass der klassische Solver viel besser abschneidet als der Pyclaw-Solver. Deshalb werden vier verschiedene tiefe neuronale Netze für den ML-basierten Solver vorgeschlagen. Die Ergebnisse zeigen, dass zwei dieser Ansätze zufriedenstellende Lösungen liefern können.

\newpage
\tableofcontents

\newpage
\begin{center}
\section*{Acknowledgement}
\end{center}
First, I would like to express my sincere thanks to my supervisors and advisor: Prof. Dr. Torsten Hoefler, Langwen Huang, for your support not only on the project advises also the life at ETHz over the past months. You lead me into the world of geophyysical fluid and deep learning. Though there are still many largely unexplored areas and uncertainties during this project, you give me inspiring thoughts during meetings and even practical engineering guidance. This thesis would be an impossible task without your support.

I would like to thank the Prof. Dr. Jörg Liesen at TUB, for the suggestions and corrections of thesis form.

I would like to thank the Prof. Dr. Siddhartha Mishra at ETHz as well, for sharing his idea to improve our work.

I would like to express my thanks to all my dear friends, for their help and support that led me through my hardest times at TUB and ETH. 

Last but not least, I am grateful to my parents, for your unconditional love and support. I would never have become the person I am today without you. You support me to study abroad and give me the warmest harbour whenever I need.

\newpage
\section*{Notation and Abbreviations}
\begin{tabular}{cc}
$\rho$ & density\\
$p$  & pressure\\
$E$  & energy\\
$h(x, t)$ & water height\\
$z(x, y)$ & water bed\\
$q$ & quantity vector\\
$g$ & gravity constant 9.8\\
$\omega$ & earth rotation rate $7.292\times 10^{-5}$\\
$a$ & radius of the earth $6.37122\times 10^{6}$\\
$S(x, q)$ & source term\\
$L(q)$ & spatial derivative(s)\\
$Q$ & numerical approximation to solution q\\
$F$ & flux function\\
$R$ & rotation matrix\\
$\hat{F}$ & numerical flux scheme \\
$\bar{F}$ & numerical approximation to solution flux\\
$\vec{x}$ & Cartesian coordinates vector\\
$\vec{u}$ & Cartesian velocity components vector\\
$x, y, z$ & Cartesian coordinates\\
$u(x, t), v(y, t), w(z, t)$ & Cartesian velocity components\\
$t$ & time\\
$\Delta T$ & time-step\\
$\Delta c$ & area of the cell\\
$\Omega$ & volume of domain\\
$\partial \Omega$ & boundary of $\Omega$\\
RK method & Runge–Kutta method\\
ML & machine learning\\
NN & neural network\\
CFD & computaional fluid dynamics\\
CNN & convolutional neural network\\
FVM & finite volume method\\
SWE & shallow water equation\\
PBC & periodic boundary conditions\\
TVD & total variation diminishing\\
ReLU & rectified linear unit\\
DACE & Data-Centric Parallel Programming\\
RMSE & root-mean-squared error\\
\end{tabular}

\newpage
\section{Introduction}
The simulation of complex physical systems described by nonlinear partial differential equations (PDEs) occupies an important place in engineering and physical sciences. Shallow water equations can be used to model fluid flow in the ocean and atmosphere~\cite{ref1}. Models of such systems can predict areas that will eventually be affected by pollution, coastal erosion, and melting of the polar ice caps. Euler's equations can be applied in many disciplines, including gas dynamics, fluid mechanics, and the environment~\cite{ref2}.

Despite the sophistication of classical solvers, direct numerical simulations at the scale required for these problems are impossible when simulating climate, weather, and tsunamis. The traditional approach uses a coarser mesh, but accuracy is sacrificed. Although still in its infancy, the application of machine learning to accelerate CFD problems is receiving more and more attention. A machine learning framework to accelerate the numerical computation of time-dependent ODEs and PDEs was proposed (Mishra, 2018)~\cite{ref3}; Bar-Sinai et al. (2019) solved burger's equation using learning data-driven discrete methods based on FVM and experimentally showed that scaling 4-8 times is remarkably accurate~\cite{ref4}. Magiera et al., (2020) applied constraint-aware neural networks to Riemann Problems~\cite{ref5}.

Maddu(2021) proposed the STENCIL-NET architecture for the data-driven, adaptive discretization of unknown nonlinear PDEs~\cite{ref6}, which is a WENO~\cite{ref7}-like solution-adaptive numerical scheme. A deep convolutional neural network (CNN) be used to forecast several essential atmospheric variables on a global grid(Weyn et al., 2020)~\cite{ref8}. Kochkov et al. (2021) obtained a 80-fold improvement in computational time by scaling the grid by a factor of ten in two-dimensional turbulence~\cite{ref9}. 

Up to our knowledge nothing has been done on the embedded data-driven shallow water equation solvers.
The rest of the thesis is structured as follows. Chapter 2 introduces Euler equation \& SWE and describes the FVM discretization method. Then we briefly introduce the data-driven method. Furthermore, a spherical grid needed in the subsequent sections is presented. In Chapters 3 and 4, we develop classical solvers for the shallow water equation and the Euler equation in the torch and dace framework and compare them with the Riemann problem solver in Pyclaw. In Chapter 5, we implemented the ml-based SWE solver using CNN using four methods. The technical details of the ML approach, such as the NN structure, are given in Chapter 2. Moreover, the total energy and potential enstrophy analysis are used for each method. Finally, chapter 6 is dedicated to the concluding discussion and future work.

\section{Theoretical Background}
The following chapter presents the theoretical background needed to develop the numerical scheme in later experiments, divided into five sections. We briefly introduce the shallow water equation and the Euler equation in the first and second parts. They exist to appear as intermittent phenomena like shock waves of the Riemann problem, whose solution is closely related to hyperbolicity.
The third part contains a brief derivation and algorithm of the finite volume method for hyperbolic equations and some numerical schemes used in the following experiments. The fourth part describes the mesh used in the spherical shallow water equation solver. In the last subsection, we briefly introduce the data-driven method and how it can be applied to PDE solving.

\subsection{The Shallow Water Equations}

When the vertical velocity and displacement at the water surface are sufficiently small compared to the wave frequency, the governing equations for shallow water flow can be obtained by depth integration of the Navier-Stokes equations, assuming hydrostatic pressure distribution and uniform distribution of horizontal velocity in the vertical direction and low bottom slope. Neglecting the diffusion terms due to viscosity and turbulence, the Coriolis terms, and the surface shear stress terms, the SWEs can be expressed in vector form~~\cite{ref10}:
\begin{equation}
\frac{\partial Q}{\partial t}+\frac{\partial F}{\partial x}+\frac{\partial G}{\partial y}=S,
\end{equation}

where $t$ denotes time; $x$ and $y$ are Cartesian coordinates on the horizontal plane; $Q$ denotes the conservation variable vector; $F$ and $G$ are the convective flux vectors in the $x$ and $y$ directions, respectively; $S$ denotes the source term, which can be further divided into a bed slope source term $S_{b}$ and a friction source term $S_{f}$ (2.5).

We consider the situation of a single space dimension first. Let $h(x, t)$ be the water level (height) of a river and $u(x, t)$ the mean velocity. Let the bottom of the river be planar. The conservation law gives the one-dimensional Shallow Water Equations:
\begin{equation}
\frac{\partial}{\partial t}\left(\begin{array}{c}
h \\
h u 
\end{array}\right)+\frac{\partial}{\partial x}\left(\begin{array}{c}
h u \\
h u^2+\frac{1}{2} g h^2
\end{array}\right)=\left(\begin{array}{l}
0 \\
0
\end{array}\right),
\end{equation}

where $g$ is the gravitation constant. This system reflects the conservation of mass and momentum, written in dependence on the water level. If the bottom of the river exhibits a profile given by a smooth function $S(x)$, then it follows a conservation law with a source term
$$\frac{\partial}{\partial t}\left(\begin{array}{c}
h \\
h u
\end{array}\right)+\frac{\partial}{\partial x}\left(\begin{array}{c}
h u \\
h u^2+\frac{1}{2} g h^2
\end{array}\right)=\left(\begin{array}{c}
0 \\
-g h S^{\prime}(x)
\end{array}\right).
$$
For a planar river bottom, it holds $S(x) \equiv$ const. And thus the above conservation law is recovered.

In case of two space dimensions, the river bottom is specified by a profile $S(x, y)$ and $h(x, y, t)$ represents the water level. Let $u(x, y, t)$ and $v(x, y, t)$ be the mean velocities in the directions $x$ and $y$, respectively. We obtain a conservation law for the two-dimensional Shallow Water Equations, with source term
\begin{equation}
\frac{\partial}{\partial t}\left(\begin{array}{c}
h \\
h u \\
h v
\end{array}\right)+\frac{\partial}{\partial x}\left(\begin{array}{c}
h u \\
h u^2 +\frac{1}{2} g h^2 \\
h u v
\end{array}\right)+\frac{\partial}{\partial y}\left(\begin{array}{c}
h v\\
h u v \\
h v^2 +\frac{1}{2} g h^2
\end{array}\right)=\left(\begin{array}{c}
0 \\
-g h \frac{\partial z}{\partial x}(x, y) \\
-g h \frac{\partial z}{\partial y}(x, y)
\end{array}\right),
\end{equation}
where two flux functions are included. The system is symmetric in $x$ and $y$. For a planar river bottom $S(x, y) \equiv$ const., it follows a system of conservation laws. It can be shown that the systems are hyperbolic.
If a friction source term $S_{f}$ is considered, the source term will be denoted as follows:
$$S=S_b+S_f=\left(\begin{array}{c}
0 \\
-g h \frac{\partial z}{\partial x}(x, y) \\
-g h \frac{\partial z}{\partial y}(x, y)
\end{array}\right)+\left(\begin{array}{c}
0 \\
-C_f u \sqrt{u^2+v^2} \\
-C_f v \sqrt{u^2+v^2}
\end{array}\right),
$$where $C_f$ is the bed roughness coefficient evaluated by
$C_f=g n^2 / h^{1 / 3},$ $n$ denotes the Manning's Roughness coefficient.

\subsection{The Euler Equations}
The Euler equations describe the dynamics of inviscid and compressible flows, and it is arguably the most famous example of nonlinear hyperbolic conservation laws. Euler equations develop based on the conservation of mass, linear momentum, and total energy. We are essentially dealing with Newton’s second law of motion and the first law of thermodynamics here. Let us consider the fluid occupying a fixed one-dimensional case and denote its density, velocity, and specific internal energy by $\rho$, $u$, and $E$, respectively. Note that these quantities describe the fluid at a fixed spatial coordinate and fixed time. If $u$ does not depend only on the (mass) density but also on other quantities, then other conserved quantities must be added to obtain a system with as many unknowns as equations. For example, we arrange~\cite{ref11},

$$
\begin{aligned}
(\rho u)_t+\left(\rho u^2+p\right)_x &=0 & & \text { (conservation of momentum), } \\
E_t+(u(E+p))_x &=0 & & \text { (conservation of energy), }
\end{aligned}
$$

with the momentum density $\rho u$ and the pressure $p$. The pressure has to be specified as a function of $\rho, \rho u, E$ according to the physical laws of gas dynamics. For example, an ideal gas satisfies

\begin{equation}
p=(\gamma-1)\left(E-\frac{1}{2} \rho u^2\right),
\end{equation}
with a constant $\gamma \in \mathbb{R}$ like $\gamma=\frac{5}{3}$.
We obtain the Euler equations of gas dynamics, which represent a system of three conservation laws. The conserved quantities are
$$q(x, t)=\left(\begin{array}{l}
q_1 \\
q_2 \\
q_3
\end{array}\right):=\left(\begin{array}{c}
\rho \\
\rho u \\
E
\end{array}\right).
$$
If we ignore the source term. The one-dimensional Euler Equations are given by the conservation law,
\begin{equation}
\frac{\partial}{\partial t}\left(\begin{array}{c}
\rho \\
\rho u \\
E
\end{array}\right)+\frac{\partial}{\partial x}\left(\begin{array}{c}
\rho u \\
\rho u^2 + p \\
u(E+p)
\end{array}\right)=\left(\begin{array}{c}
0 \\
0 \\
0 \\
\end{array}\right).
\end{equation}
For the pressure, it holds $p\left(q_1, q_2, q_3\right)=(\gamma-1)\left(q_3-\frac{1}{2} q_2^2 / q_1\right)$, in case of $(2.4)$.

\subsection{Finite Volume Method for hyperbolic equations}
In this chapter, we will introduce efficient schemes for the scalar conservation law. Central differences cannot be used to approximate the conservation law, even in the simplest case of linear transport. The crucial step in designing an efficient scheme for linear transport equations was to upwind it by taking derivatives in the direction of information propagation. For a linear equation with constant coefficients like $q_t+aq_x=0$, the direction of information propagation is given by the constant velocity field. For a nonlinear conservation law like $q_t+F(q)_x=0$, the wave speeds depend on the solution itself and can not be determined a priori. Thus, it is not clear how differences can be upwind.
The key idea underlying finite difference schemes is to replace the derivatives in equations like $q_t+F(q)_x=0$ with a finite difference. This procedure requires the solutions to be smooth and the equation to be satisfied point-wise~\cite{ref2}. However, the solutions to the scalar conservation law are not necessarily smooth, so the Taylor expansion essential for replacing derivatives with finite differences is no longer valid. Hence, the finite difference framework is not suited for approximating conservation laws. Instead, we need to develop a new paradigm for designing numerical schemes for scalar conservation laws.
\subsubsection{Discrete Consrvation}
The first step in any numerical approximation is to discretize the computational domain in both space and time.
\paragraph{The grid}~\\
For simplicity, we consider a uniform discretization of the domain $\left[x_L, x_R\right]$. The discrete points are denoted as $x_i=x_L+(i+1 / 2) \Delta x$ for $i=0, \ldots, N$, where $\Delta x=\frac{x_R-x_L}{N+1}$. We also define the midpoint values
$$
x_{i-1 / 2}=x_i-\Delta x / 2=x_L+i \Delta x.
$$
For $i=0, \ldots, N+1 $. These values define computational cells or control volumes
$$
\Omega_i=\left[x_{i-1 / 2}, x_{i+1 / 2}\right) .
$$
As we will see soon, the finite volume method uses the control volumes $\Omega_i$ instead of the mesh points $x_i$. We use a uniform discretization in time with time step $\Delta t$. The time levels are denoted by $t^n=n \Delta t$. See Figure 1 for an illustration of the grid.
\begin{figure}
\centering
\includegraphics[width=0.7\textwidth]{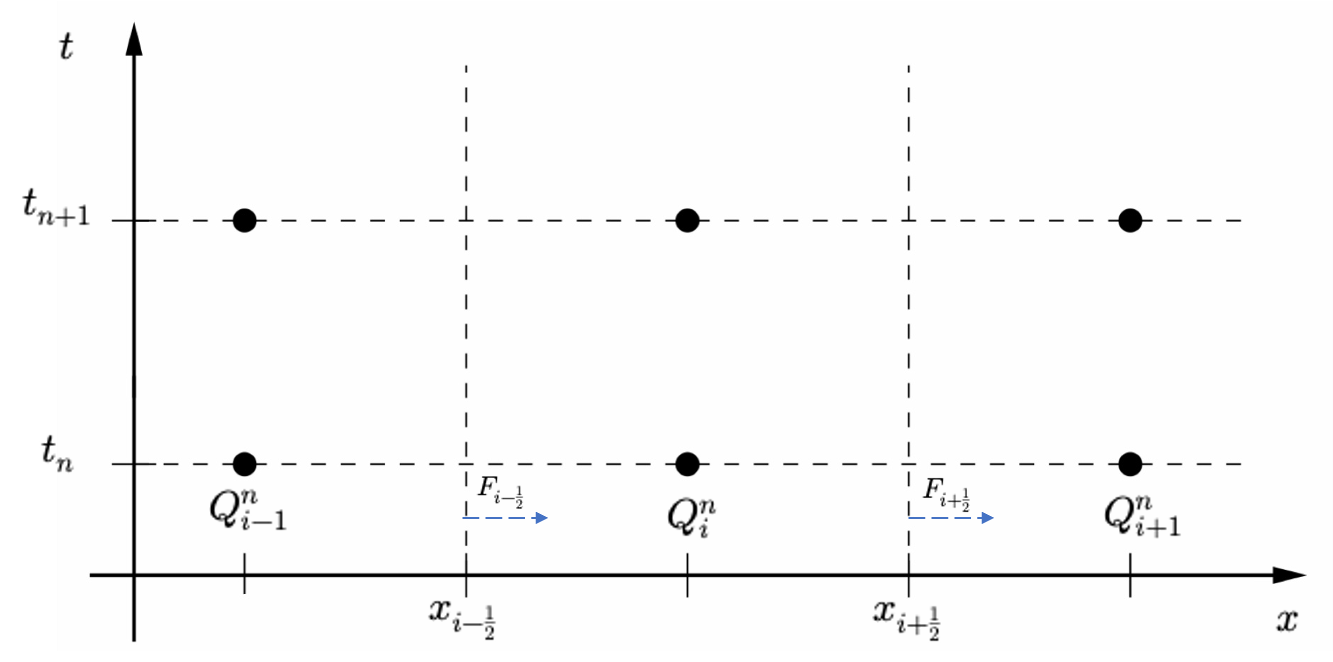}
\caption{\label{fig:FVM_grid}A typical finite volume grid displaying cell averages and fluxes.}
\end{figure}

\paragraph{Cell averages}~\\
A finite difference method is based on approximating the point values of the solution of a PDE. This approach is not suitable for conservation laws as the solutions are not continuous and point values may not make sense. Instead, we change the perspective and use the cell averages:
\begin{equation}
Q_i^n \approx \frac{1}{\Delta x} \int_{x_{i-1 / 2}}^{x_{i+1 / 2}} q\left(x, t^n\right) dx,
\end{equation}

at each time level $t^n$ as the main object of interest for our approximation.
The cell average is well defined for any integrable function, hence also for the solutions of the conservation law $q_t+F(q)_x=0$. The aim of the finite volume method is to update the cell average of the unknown at every time step.
\paragraph{Integral form of the conservation law}~\\
Integral form of the conservation law. Assume that the cell averages $Q_i^n$ at some time level $t^n$ are known. How do we obtain the cell averages $Q_i^{n+1}$ at the next time level $t^{n+1}$. A finite volume method computes the cell average at the next time level by integrating the conservation law $q_t+F(q)_x=0$ over the domain $\left[x_{i-1 / 2}, x_{i+1 / 2}\right) \times\left[t^n, t^{n+1}\right)$. This gives
$$
\int_{t^n}^{t^{n+1}} \int_{x_{i-1 / 2}}^{x_{i+1 / 2}} q_t d x d t+\int_{t^n}^{t^{n+1}} \int_{x_{i-1 / 2}}^{x_{i+1 / 2}} F(q)_x d x d t=0.
$$

Using the fundamental theorem of calculus gives
\begin{equation}
\begin{aligned}
&\int_{x_{i-1 / 2}}^{x_{i+1 / 2}} q\left(x, t^{n+1}\right) dx-\int_{x_{i-1 / 2}}^{x_{i+1 / 2}} q\left(x, t^n\right) dx \\
&=-\int_{t^n}^{t^{n+1}} F\left(q\left(x_{i+1 / 2}, t\right)\right) d t+\int_{t^n}^{t^{n+1}} F\left(q\left(x_{i-1 / 2}, t\right)\right) dt.
\end{aligned}
\end{equation}
Defining
\begin{equation}
\bar{F}_{i+1 / 2}^n=\frac{1}{\Delta t} \int_{t^n}^{t^{n+1}} F\left(q\left(x_{i+1 / 2}, t\right) d t\right.
\end{equation}
and dividing both sides of (2.7) by $\Delta x$, we obtain
\begin{equation}
Q_i^{n+1}=Q_i^n-\frac{\Delta t}{\Delta x}\left(\bar{F}_{i+1 / 2}^n-\bar{F}_{i-1 / 2}^n\right).
\end{equation}
Equation (2.9) is a statement of conservation: The rate of change of the cell average is given by the difference in fluxes across the boundary of the cell. See Figure 2 for an illustration. Note that the relation (2.9) is not explicit as $\bar{F}$ need a priori knowledge of the exact solution. The main ingredient in a finite volume scheme is a clever procedure to approximate these fluxes.

\paragraph{CFL condition}~\\
The Courant-Friedrichs-Lewy or CFL condition is a stability condition for unsteady numerical methods that simulate convection or wave phenomena. We consider a linear equation with constant coefficients like $q_t+aq_x=0$, then the Courant number is defined as:
$$C=\frac{a \Delta t}{\Delta x},$$
where $a$ is the vekicuty magnitude, $\Delta t$ is the time step and $\Delta x$ is the length between mesh cells
Courant–Friedrichs–Lewy condition takes its name after Richard Courant, Kurt Friedrichs, and Hans Lewy  who introduced it in their paper~\cite{ref12}.
The condition is expressed in terms of the Courant number as 
\begin{equation}
C=\frac{a \Delta t}{\Delta x} \leqslant 1.
\end{equation}
Once we have determined the CFL number, we can calculate the required $\Delta t$ at each iteration. In the subsequent experiments in Chapters 3,4,5, we pre-calculate the $\Delta t$ under CFL number=0.3 with $\Delta x$ equal to 128 and then calculate the corresponding $\Delta t$ according to the resolution ratio.

\begin{figure}
\centering
\includegraphics[width=0.7\textwidth]{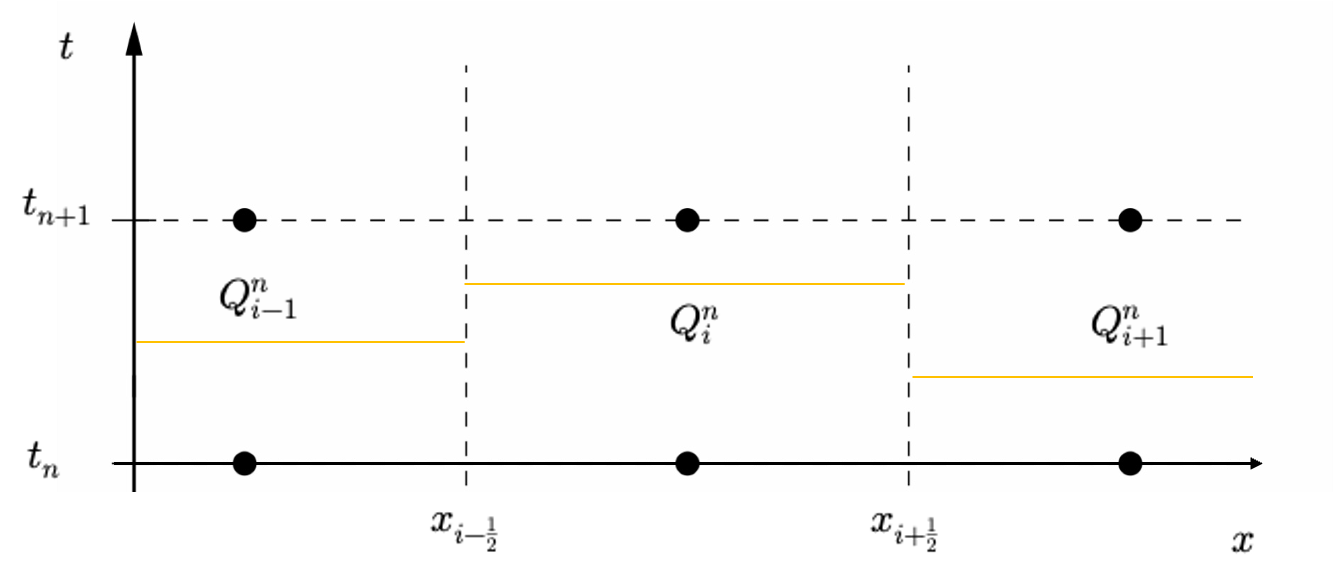}
\caption{\label{fig:FVM_grid}Cell averages define Riemann problems at every interface.}
\end{figure}

\subsubsection{Numerical Flux for FVM schemes}
In order to approximate the numerical solution $Q$ of the the conservation law $q_t+F(q)_x=0$, we should use numerical flux schemes $\hat{F}$ to get the numerical flux $\bar{F}$ as in (2.9).
\paragraph{Lax-Friedrichs flux}~\\
The Lax-Friedrichs flux~\cite{ref13} is a simple numerical flux. Only the $F$ should be know to compute this flux:
\begin{equation}
\bar{F}_{i+1 / 2}^n=F_{i+1 / 2}^{LF}=\frac{1}{2}\left(F(Q_i^n)+F(Q_{i+1}^n)\right)-\frac{\Delta x}{2 \Delta t}\left(Q_{i+1}^n-Q_i^n\right).
\end{equation}

\paragraph{Rusanov flux}~\\
The Rusanov flux~\cite{ref14} computes the cell-interface fluxes irrespective of the direction of information propagation as:
\begin{equation}
\bar{F}_{i+1 / 2}^n =F_{i+1 / 2}^{Rusanov}=\frac{1}{2}\left(F(Q_i^n)+F(Q_{i+1}^n)\right)-s_R(Q_{i+1}^n-Q_i^n),
\end{equation}
where
$$s_R = \frac{\max \left(\left|F^{\prime}\left(Q_i^n\right)\right|,\left|F^{\prime}\left(Q_{i+1}^n\right)\right|\right)}{2}.$$
\paragraph{Roe flux}~\\
A common method for solving nonlinear equations is to linearize them. Linearization entails replacing the nonlinear flux function $q_t+F(q)_x=0$ with a locally linearized version
$$F(q)_x=F^{\prime}(Q) Q_x \approx \hat{A}Q_x,$$
where $\hat{A} \approx f^{\prime}$ is a constant state around which the nonlinear flux function is linearized. There are many ways in which a $\hat{A}(q_i, q_{i+1})$ can be constructed for the linearizing state, which should follow a series of properties and then find a way to construct it.
\begin{enumerate}
    \item $\hat{A}$ should have purely real eigenvalues $\hat{\lambda}_i\left(q_i, q_{i+1}\right)$ with corresponding set of complete eigenvectors $\hat{v}_i$.
    \item Consistency with the exact Jacobian: $\hat{A}(q, q)=A(q)$.
    \item Conservation: $f\left(q_{i+1}\right)-f\left(q_i\right)=\hat{A}\left(q_i, q_{i+1}\right)\left(q_{i+1}-q_i\right)$
\end{enumerate}

\begin{equation}
\bar{F}_{i+1 / 2}^n=F_{i+1 / 2}^{Roe}=\frac{1}{2}\left(F(Q_i^n)+F(Q_{i+1}^n)\right)-\frac{1}{2}\sum_{p=1}^n\left|\tilde{\lambda}^{(p)}\right| \tilde{\alpha}^{(p)} \tilde{r}^{(p)},
\end{equation}
where
$$\tilde{\alpha}^{(p)}=\tilde{l}^{(p)}\left(q_{i+1}-q_i\right),$$
$\tilde{\lambda}^{(p)}$, $\tilde{r}^{(p)}$, $\tilde{l}^{(p)}$ being the eigenvalues and right and left eigenvectors of  $\hat{A}$ , respectively ($p$ extends from  1 to the number of equations $n$ of the system). 
This can also written by
\begin{equation}
\bar{F}_{i+1 / 2}^n=F_{i+1 / 2}^{Roe}=\frac{1}{2}\left(F(Q_i^n)+F(Q_{i+1}^n)\right)-\frac{1}{2}\bar{R}|\bar{\Lambda}| \bar{R}^{-1}\left(Q_{i+1}^n-Q_i^n\right),
\end{equation}
where $\bar{R}^{-1}$ and $\bar{R}$ are the matrices diagonalizing the Jacobian $\hat{A}$, $\Lambda$ is the diagonalized Jacobian and $|\hat{\Lambda}|$ is the matrix which has on the diagonal the relative values of the eigenvalues of $\hat{A}$.
The finite volume scheme with the Roe flux $(2.13)$ or $(2.14)$ is termed the Roe or Murman-Roe scheme~\cite{ref15}. 

\paragraph{HLL flux}~\\
The HLL scheme~\cite{ref16} computes the cell-interface fluxes based on left and right side fluxes as well as on the direction of information propagation. $F_L=F(Q_i^n)$ and $F_R=F(Q_{i+1}^n)$ are the fluxes corresponding to the left-side and right-side of the cell interface, and $s_L$ and $s_R$ describe the slowest and fastest minimum propagation velocities in the vicinity of the cell interface computed:
$$\left\{\begin{array}{l}
s_R=\max \left(\lambda_L, \lambda_R\right) ,\\
s_L=\min \left(\lambda_L, \lambda_R\right).
\end{array}\right.$$
The HLL solver is defined:
\begin{equation}
\bar{F}_{i+1 / 2}^n=\left\{\begin{array}{cl}
F_L & \text { if } 0 \leq s_L ,\\
F_{i+1 / 2}^{Hll}=\frac{s_R F_L-s_L F_R+s_L s_R\left(Q_{i+1}^n-Q_{i}^n\right)}{s_R-s_L} & \text { if } s_L<0 \leq s_R, \\
F_R & \text { if } s_R<0.
\end{array}\right.
\end{equation}

\paragraph{HLLE flux}~\\
The HLLE~\cite{ref17} is similar to HLL scheme:
\begin{equation}
\bar{F}_{i+1 / 2}^n=\left\{\begin{array}{cl}
F_L & \text { if } 0 \leq s_L, \\
F_{i+1 / 2}^{HllE}=\frac{s_R F_L-s_L F_R+s_L s_R\left(Q_{i+1}^n-Q_{i}^n\right)}{s_R-s_L} & \text { if } s_L<0 \leq s_R, \\
F_R & \text { if } s_R<0,
\end{array}\right.
\end{equation}
but $s_L$ and $s_R$ defined by:
$$\left\{\begin{array}{l}
s_R=\max \left(\lambda_R, \lambda_m, 0\right) ,\\
s_L=\min \left(\lambda_L, \lambda_m, 0\right),
\end{array}\right.$$
where $\lambda_m$ is computed by the Roe average state $q_{m}$. The HLLE solver is extremely robust as it is both positivity preserving and satisfies and entropy inequality automatically. However, due to the single intermediate state approximation, the HLLE solver cannot resolve non-fast isolated discontinuities, making it quite dissipative.
\paragraph{HLLC flux}~\\
The HLLC~\cite{ref16} is a modification of the HLL Riemann solver.
\begin{equation}
\bar{F}_{i+1 / 2}^n=F_{i+1/2}^{HLLC}= \begin{cases}F_L & \text { if } 0 \leq s_L, \\ F_{* L}=F_L+s_L\left(Q_{* L}-Q_{i}^n\right) & \text { if } s_L \leq 0 \leq s_* ,\\ F_{* R}=F_R+s_R\left(Q_{* R}-Q_{i+1}^n\right) & \text { if } s_* \leq 0 \leq s_{R} ,\\ F_R & \text { if } 0 \geq s_R.\end{cases}
\end{equation}
We have four unknown vectors: $Q_{*_L}$,  $Q_{*_R}$, $F_{*_L}$, $F_{*_R}$. We first solve for the states: $Q_{*_L}$, $Q_{*_R}$, then solve for the fluxes: $F_{*_L}$, $F_{* R}$. There are several mechanisms to define the propagation velocity of
the intermediate state, and the choice of propagation velocity can affect the performance of the HLLC solver. Furthermore, different equations have different approximations~\cite {ref18,ref19}.

\subsubsection{Second order Schemes}
\paragraph{Spatial discretization}~\\
Here we use slop limiter~\cite{ref20} to reconstruct the left and right state variables, which plays a vital role in suppressing spurious numerical oscillations. They can be written in different equivalent forms and may be equivalent to the corresponding flux limiters. Most are derived strictly for a single nonlinear conservation equation or constant coefficient equations in one-dimensional space. In later experiments, we mainly use the MINMOD limiter, originally introduced as a TVD flux limiter. The MINMOD limiter can be employed to reconstruct a second-order scheme as follows:
\begin{equation}
\left\{\begin{array}{l}
Q_{i+1 / 2}^L=Q_i+ MINMOD \left(\frac{Q_i-Q_{i-1}}{\Delta x}, \frac{Q_{i+1}-Q_i}{\Delta x}\right) * \frac{\Delta x}{2}  ,\\
Q_{i+1 / 2}^R=Q_{i+1} + MINMOD \left(\frac{Q_{i+1}-Q_i}{\Delta x}, \frac{Q_{i+2}-Q_{i+1}}{\Delta x}\right) * \frac{\Delta x}{2} ,
\end{array}\right.
\end{equation}
where,
$$
MINMOD(a, b)=\left\{\begin{array}{l}
a \text { If }|a|<|b| \text { and } a b>0 ,\\
b \text { If }|b|<|b| \text { and } a b>0 ,\\
0 \text { If } a b \leq 0.
\end{array}\right.
$$
We refer to it as the MINMOD scheme.
\paragraph{Temporal discretization}~\\
For the temporal discretization, We could use the classic Heun's method~\cite{ref21}, where $L(q)$ is the spatial derivative(s)
\begin{equation}
\begin{aligned}
&q^{(1)}=q^n+\Delta t L\left(q^n\right),\\
&q^{n+1}=\frac{1}{2} q^n+\frac{1}{2} q^{(1)}+\frac{1}{2} \Delta t L\left(q^{(1)}\right).
\end{aligned}
\end{equation}
To maintain the TVD(total variation diminishing) property $T V\left(u^{n+1}\right) \leq T V\left(u^n\right)=\sum_j\left|u_{j+1}-u_j\right|$ while achieving higher order accuracy in time, we should use the high order TVD (total variation diminishing) Runge-Kutta method. Fourth order TVD RK method~\cite{ref22} is presented:
\begin{equation}
\begin{aligned}
q^{(1)} &=q^n+\Delta t L\left(q^n\right) ,\\
q^{(2)} &=\frac{3}{4} q^n+\frac{1}{4} q^{(1)}+\frac{1}{4} \Delta t L\left(q^{(1)}\right) ,\\
q^{n+1} &=\frac{1}{3} q^n+\frac{2}{3} q^{(2)}+\frac{2}{3} \Delta t L\left(q^{(2)}\right).
\end{aligned}
\end{equation}

\subsubsection{The algorithm for hyperbolic equations}
We can consider the general conservation law problem, represented by the following PDE,
\begin{equation}
\partial_t q+\nabla \cdot F(q)=0.
\end{equation}
Then finite volume method can be applied on a control volume $\Omega$ using the integral form of the conservation law,
\begin{equation}
{\partial_t} q+\frac{1}{\Delta c} \oint_{\partial \Omega} F(q) \cdot \vec{n} d s=0,
\end{equation}
where $\vec{n}$ is the outward pointing unit normal vector at the boundary $\partial \Omega$ and $F(q) \cdot \vec{n}$ is the flux normal to the boundary. It leads to a discrete finite volume method (2.9) of the form,
\begin{equation}
Q^{n+1}=Q^n-\frac{\Delta t}{\Delta c} \sum_{i=1}^N h_i \bar{F}_i^n,
\end{equation}
where $\bar{F}_i^n$ represents a numerical approximation to the average normal flux across the $i$-th side of the grid cell, $N$ is the number of sides. In the 2D case, $h_i$ is the length of the $i$-th side, and $\Delta c$ is the area measured in physical space. In the 3D case, $h_i$ is the area of the cell interface, and $\Delta c$ will be the volume of the cell in physical space. We could now apply any unstructured grid method to approximate hyperbolic problems on the grids discussed in this theis.

We will use it for two-dimensional planar grids and the logically rectangular mapped grids in the following chapter. The physical space could be any quadrilateral domain for the logically rectangular grids, but the computational space is a square. Section 3.3 shows how the method can approximate SWEs equations on the sphere.
In the next step, we could divide domain $\Omega$ into many small cells $\omega$, and each cell also satisfies (2.21), then we have:
$$\int_{\omega}\left(\partial_t q+{\nabla} F(q)\right)dA=0,$$
similar to (2.22),
$$\frac{\partial}{\partial t} q+\frac{1}{\Delta c} \oint_{\partial \omega} F(q) \cdot \vec{n} d s=0.$$
Approximations to the cell intercell fluxes are obtained by solving Riemann problems for a one-dimensional system of equations normal to the interface. For isentropic equations such as the Euler, the shallow water, or the acoustics equations, the flux $\bar{F}$ at a grid cell interface can be calculated by first rotating the velocity components of the cell average values of the quantity $q$ on both sides of the interface into the direction normal and tangential to the interface. A one-dimensional Riemann problem is solved with these modified data, and flux is calculated. Finally, the momentum components of the flux are rotated back to obtain the flux in physical space.

To be more specific, we now consider the $\vec{n}=(n^{1}, n^{2})$ and $\vec{t}=(t^{1}, t^{2})$ are the normal vector and tangent vector of a cell interface respectively. Then the rotation matrix $R$, which rotates the velocity components (e.g., for the shallow water, where $q = \left( h,\  hu,\  hv\right)^{T}$ has the form,

\begin{equation}
q^{\bullet} = R\cdot q = \left( \begin{array}{ccc}1&0&0\\ 0&n^{1}&n^{2}\\0&t^{1}&t^{2}\end{array} \right) \cdot \left( \begin{gathered}h\\ hu\\ hv\end{gathered} \right).
\end{equation}
To solve the hyperbolic equations, the algorithm~\cite{ref23} consists of the following steps:
\begin{enumerate}
    \item Determine left and right state ($Q_{L}^{n\bullet}$ and right $Q_{R}^{n\bullet}$) between the cell interface by rotating the velocity components, i.e. $$Q_L^{n\bullet} = R\cdot Q_L^n, Q_R^{n\bullet} = R\cdot Q_R^n.$$
    \item Computer the $\bar{F}^{n\bullet}$ using any numerical flux scheme $\hat{F}$ presented in 2.3.2$$\bar{F}^{n\bullet} =\hat{F}(Q_L^{n\bullet}, Q_R^{n\bullet}).$$
    \item Rotate numerical flux $\bar{F}^{n\bullet}$ back to Cartesian coordinates by $$\bar{F}^{n} = R^{T}\cdot \bar{F}^{n\bullet}.$$
    \item Repeat the step 1,2,3 to calculate all the sides' numerical flux $\bar{F}_i^{n}$.
    \item Update the cell state $Q^{n+1}$ using the finite volume method (2.23), note that the $h_{i}$ and $\Delta c$ could be computed in advance.
\end{enumerate}

\subsection{Logically rectangular grid on sphere}
There are many grids available for transport on a spherical manifold, and the most popular are the latitude-longitude grid (Lin \& Rood, 1996), the cubed-sphere grid (Nairetal., 2005; Putman \& Lin, 2007; Ronchietal., 1996), the Yin-Yang grid (Hall \& Nair, 2013; Kageyama \& Sato, 2004), and Logically rectangular grids(Donna \& Christiane \& Randall, 2007)~\cite{ref24}. Latitude-longitude grids are algorithmically simple, yet converging meridians near the poles lead to Courant-Friedrichs-Lewy (CFL) stability problems for Eulerian schemes, a strong polar singularity, and the need for zonally global methods to stabilize polar flow for both Eulerian and Lagrangian methods. Yin-Yang grids are orthogonal and easy to implement, and they have the added benefit over latitude-longitude grids of quasi-uniform grid spacing. But they have the problem that maintaining mass conservation at high-order accuracy on a globally rectangular grid requires the use of a global mass fixer. The cubed-sphere’s gridding system uniquely spans the sphere to allow trivial exact and local mass conservation, is logically rectangular, is quasi-uniform, and requires no global operations. Its complication is that it uses six separate nonorthogonal coordinate systems that require metric transformations, and more complicated boundary conditions should be considered. We choose the Logically rectangular grids because it maintains all four preferred properties: entirely local operations, local mass conservation, uniform Cartesian grids and a strictly logically rectangular mesh.
\subsubsection{Mappings for the disk}
In this part, we describe a mapping which maps the computational domain $[-1,1] \times[-1,1]$ to the unit disk. To describe the mapping, we focus our attention on the region of the square in which computational coordinates $(x_c, y_c)$ satisfy $|y_c| \leq x_c$. This region corresponds to the upper triangular region between the two diagonals of the square. We refer to this region as the north sector of the computational domain. The mapping in this sector is based on the idea that we can map a vertical line segment between $(d,-d)$ and $(d, d)$ is mapped to a circular arc of radius $R(d)$ that passes through the points $(D(d),-D(d))$ and $(D(d), D(d))$. The mapping is fully defined by the functions $R(d)$ and $D(d)$ as discussed further below. It is easy to compute that the center of this circle of radius $R(d)$ is at $\left(x_0, y_0\right)=\left(D(d)-\sqrt{R(d)^2-D(d)^2}, 0\right)$. The point $\left(x_c, y_c\right)$ is then mapped to

$$
\begin{aligned}
&y_p=y_c D(d) / d ,\\
&x_p=D(d)-\sqrt{R(d)^2-D(d)^2}+\sqrt{R(d)^2-y_p^2}.
\end{aligned}
$$

Similar mappings are used in the other four sectors, for example points on the horizontal line segment between $(-d, d)$ and $(d, d)$ are mapped to points on the circular arc of radius $R(d)$ passing through $(-D(d), D(d))$ and $(D(d), D(d))$.

    




\noindent There are many choices for $R(d)$ and $D(d)$ that lead to useful grids. For example, one choice that is well-suited for the unit disk is
$$\begin{aligned}
&D(d)=d / \sqrt{2} ,\\
&R(d)=1 .
\end{aligned}$$
\subsubsection{A mapping for the unit sphere}
The above mapping for the unit disk can be used directly to create a grid for the unit hemisphere or sphere. To parameterize the hemisphere, for example, we could simply describe the hemispherical surface as a function of physical locations $(x_p, y_p)$ from the disk mapping:
\begin{equation}
z_p= \sqrt{1-(x_p^2+y_p^2)}.
\end{equation}
Then the points $(x_p, y_p, z_p)$ lie on the upper hemisphere. We can apply a similar mapping to points in the computational domain $[-3,-1] \times[-1,1]$ to map these points to the lower hemisphere. The two mappings map the rectangle $[-3,1] \times[-1,1]$ to the sphere's surface. Figure 3 shows the resulting grid.


\begin{figure}[H]
\centering
\includegraphics[width=1\textwidth]{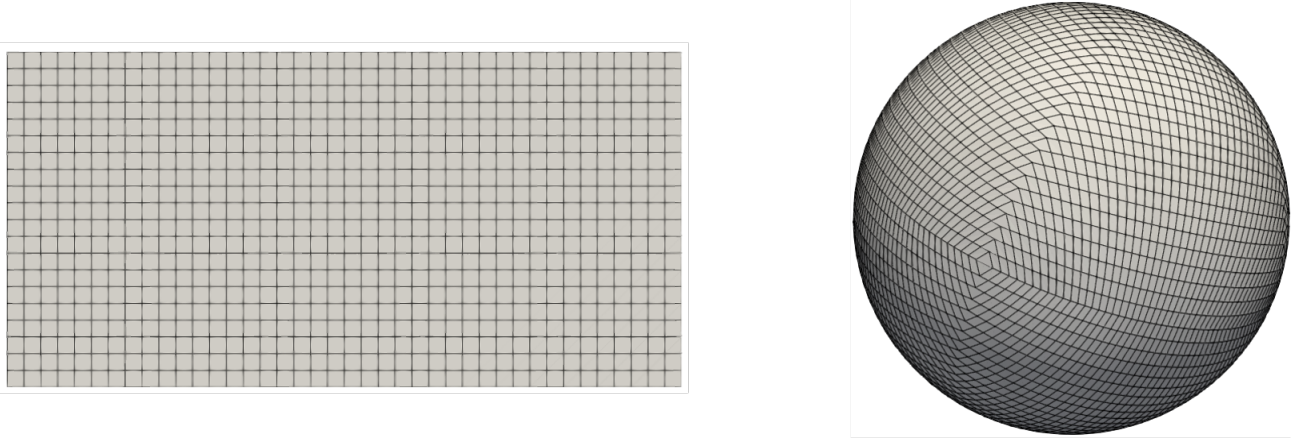}
\caption{\label{fig:sphere_grid}On the left is the computational grid and the sphere grid shown in the right picture.}
\end{figure}

\subsection{Machine Learning Accelerated CFD}
This section presents two methods for solving CFD using neural networks. The first one uses NN to generate derivatives, and the second uses NN reconstruction to get the boundary values, similar to the idea of 2.3.3.
\subsubsection{Data-driven approximation of the spatial derivative}
We consider a generic PDE, which has a continuous solution $u(x, t)$ ~\cite{ref4}:
\begin{equation}
\frac{\partial u}{\partial t}=F\left(t, x, u, \frac{\partial u}{\partial x_i}, \frac{\partial u}{\partial x_i \partial x_j}, \cdots\right),
\end{equation}
We can use traditional numerical methods for solving the problem, such as the finite differences method and finite volume method. These methods require discretizing the spatial domain and calculating the approximate spatial derivatives. There are various methods for this approximation, such as polynomial expansion and spectral dichotomy. For example, the one-dimensional finite difference Euler method:
$$\frac{\partial u(x_i,t)}{\partial x} = \frac{u_{i+1}(t) - u_{i}(t)}{\Delta x} + \mathcal{O}(\Delta x),$$
where the solution $u(x, t)$ is represented by its values at node points $u_i(t)=$ $u\left(x_i, t\right)$ and $\Delta x=x_i-x_{i-1}$ is the spatial resolution.

Similarly, we can obtain the rest of the spatial derivatives $\frac{\partial^n u(t)}{\partial x^n} \approx \sum_i \alpha_i^{(n)} u_{i}(t).$  Traditional schemes use a set of precomputed coefficients for all points in the space, and more advanced approaches alternate between different sets of coefficients according to local rules. We can use machine learning to determine these sophisticated coefficients between stencils.

Now equation(2.41) can be transformed into a set of coupled ordinary differential equations consisting of the values of each node $u_i(t)$:
\begin{equation}
\frac{\partial u_i(t)}{\partial t}=F\left(t, x, u_1(t), \cdots, u_N(t)\right),
\end{equation}
Then we can use the line method like the Euler or Runge–Kutta method to calculate the numerical solution of the equation for each spatial node.

 First, we generate a training set from high-resolution data and learn discrete approximations of the derivative from this data set. It has a trade-off with computational cost, which can be mitigated by performing high-resolution simulations on a small system, developing local approximations for solving the manifold, and using them to solve the equations for a more extensive system at a lower spatial resolution.

Our network representation is the generalized finite difference formula for the spatial derivative: the network's output is a column of coefficients $\alpha_1, ... \alpha_{N}$, such that the n-th derivative is performed as a pseudo linear formula. As a pseudo-linear filter, the coefficients $\alpha_i^{(n)}\left(u_1(t), u_2(t), \ldots\right)$ through its dependence on space and time for the field values in neighboring cells. Finding the optimal coefficients is the key to the method.

\subsubsection{Data-driven approximation of the numerical flux}
\paragraph{Calculate boundary Flux directly from CNN}~\\
This method uses CNN to directly calculate the boundaries' Flux $F_{i}^l$ and $F_{i}^r$ from the cell average states $Q_i$. Then we can update the state values using Heun's method or the TVD RK method introduced in 2.3.3. We performed a numerical experiment of this method in section 5.1.
\begin{figure}[H]
\centering
\includegraphics[width=1\textwidth]{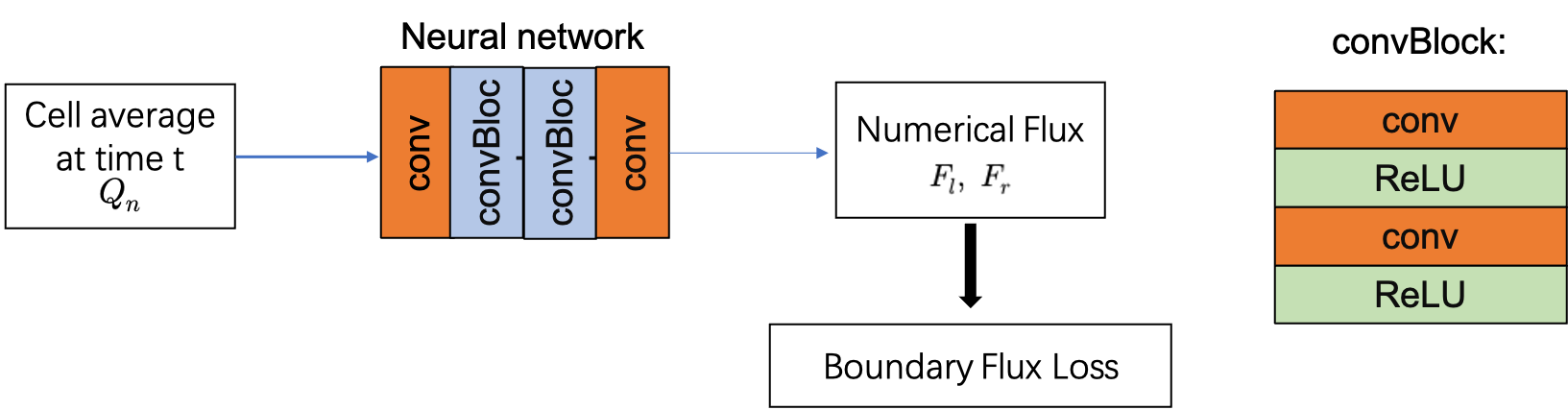}
\caption{\label{fig:nn1} First NN structure: CNN output the boundaries' numerical flux directly.}
\end{figure}

\subsubsection{Data-driven approximation of the boundary state}
In addition to using the data-driven method to get the time derivative directly, i.e., the flux value, we can also try to use the neural network to reconstruct the value of center cell values to the cell boundary values, which then we can use the numerical flux to get the high-accuracy flux. In this thesis, we propose three CNN-based reconstruction methods. These methods are similar to the principles of the high-precision Weno scheme and aim to obtain stable, non-oscillatory, and sharp discontinuity transitions. The neural network can obtain the linear weights of local regions, which theoretically leads to arbitrary order interpolation.

\paragraph{Calculate boundary states with CNN generated linear coefficients}~\\
The idea of this method is to use CNN to calculate the linear constraint coefficients interpolated to the boundary values. Then the boundary values are obtained by multiplying them with the corresponding values of the stencil. And then, we could calculate the numerical fluxes. Here, our training set is high-resolution results coarsen to the center and boundary values of the cells. Since the flux obtained by interpolation may be oscillatory, we can apply flux limiters to eliminate the discontinuities, which are presented below:
$$
\begin{aligned}
&F\left(q_{i+\frac{1}{2}}\right)=F_{i+\frac{1}{2}}^{\text {low }}-\phi\left(r_i\right)\left(F_{i+\frac{1}{2}}^{\text {low }}-F_{i+\frac{1}{2}}^{\text {high }}\right) ,\\
&F\left(q_{i-\frac{1}{2}}\right)=F_{i-\frac{1}{2}}^{\text {low }}-\phi\left(r_{i-1}\right)\left(F_{i-\frac{1}{2}}^{\text {low }}-F_{i-\frac{1}{2}}^{\text {high }}\right),
\end{aligned}
$$
where 
$F^{low}$ is the low-resolution flux computed by the low-resolution grid, $F^{high}$ is the high-resolution flux obtained by interpolated value, and $r$ represents the ratio of successive gradients on the solution mesh, i.e.,
$r_i=\frac{u_i-u_{i-1}}{u_{i+1}-u_i}$. We present the experiments with this approach in section 5.2
\begin{figure}[H]
\centering
\includegraphics[width=1\textwidth]{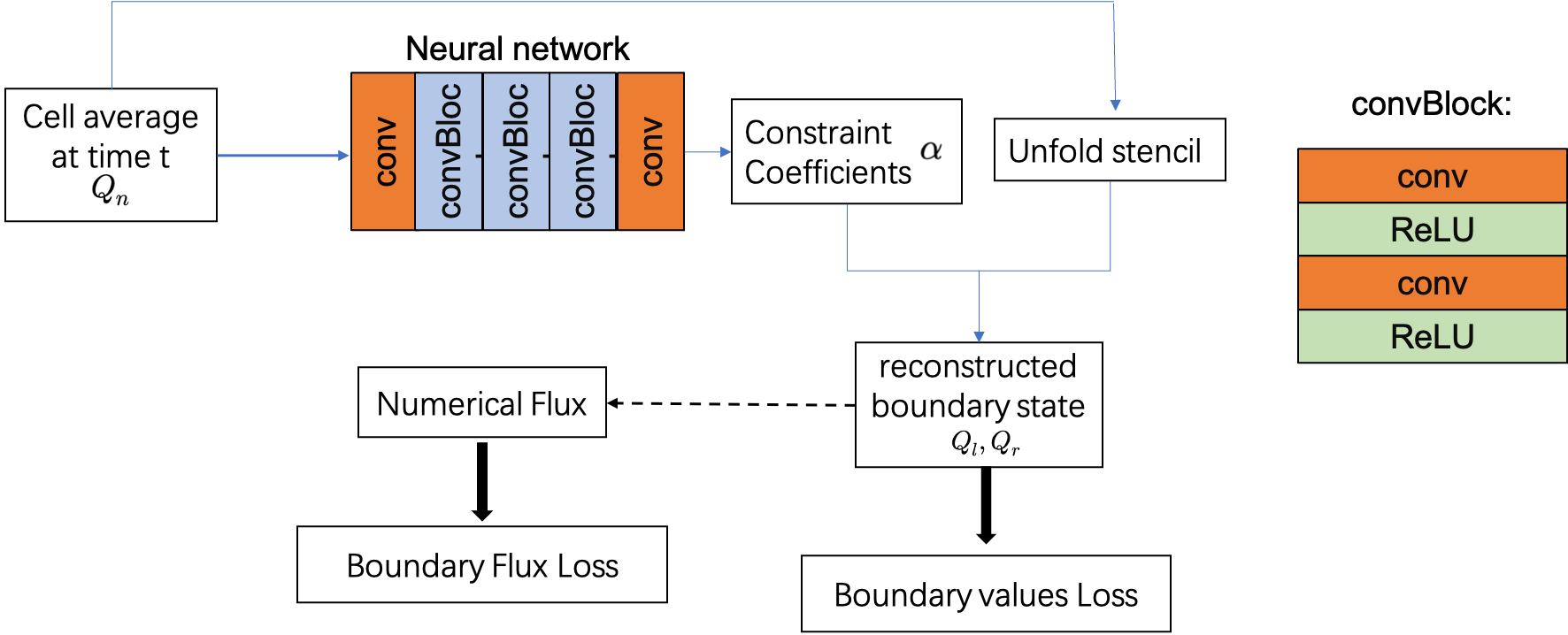}
\caption{\label{fig:nn2} Second NN structure: Calculate boundary states with CNN generated linear coefficients.}
\end{figure}

\paragraph{Calculate boundary states directly from CNN}~\\
In this approach, we use CNN to reconstruct the states at boundaries directly $Q_{i}^l$ and $Q_{i}^r$ from the cell average states $Q_i$. Here we could use the flux vector $F(q)$ directly or any numerical flux scheme. Note that here $Q_{i}^r = -Q_{i+1}^l$. For the neural network, due to the different properties of the conservation components (the height term needs to be positive when computing the numerical flux). For the input of the neural network, we first normalize the input values and split them into the height term $h$ and the momentum term $hu$. After that, we use several layers of Conv functions and one layer of GeLu as the activation function between each layer. The last layer of CNN used in the height term needs to be Softplus to ensure that the reconstruction height is greater than zero. 

\begin{figure}[H]
\centering
\includegraphics[width=1\textwidth]{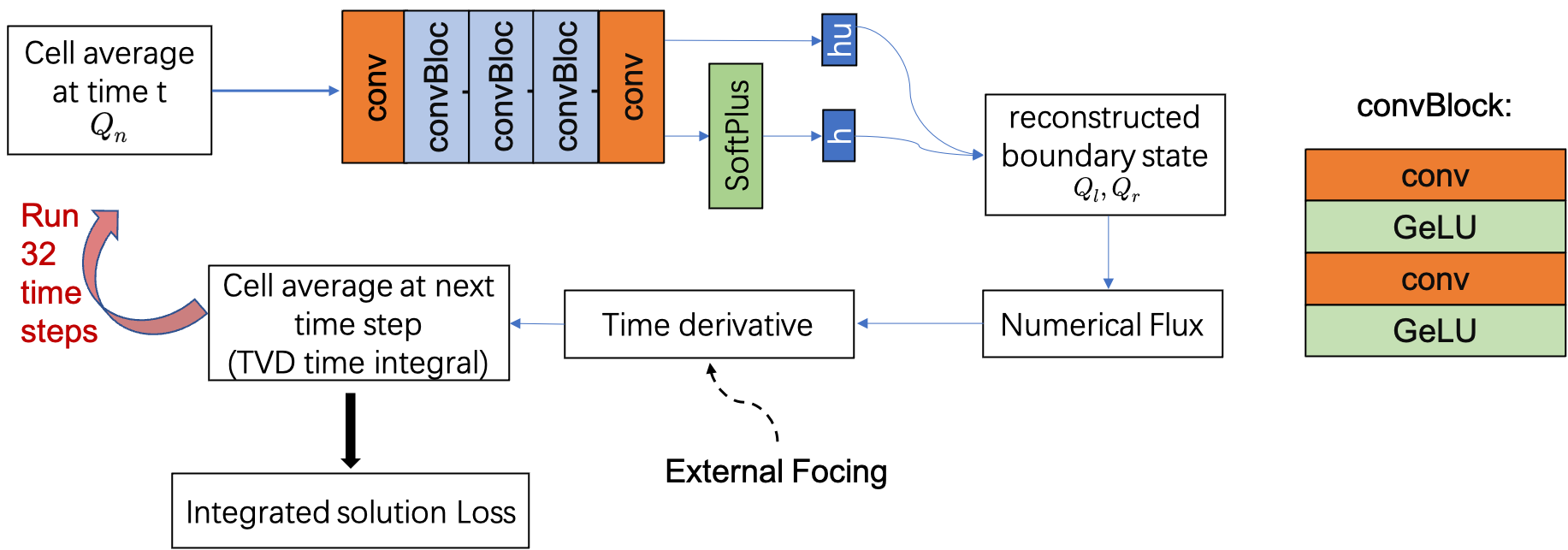}
\caption{\label{fig:nn3}Third NN structure: Calculate boundary states directly from CNN.}
\end{figure}

After obtaining the boundary state from CNN, we compute the $Q_{i}^r$ and $Q_{i}^l$ and update each cell state. Including the numerical solver in the training loss ensures full "model consistent" training, which can improve the stability of the training. In our following experiments, we use the sum of the results of 32 steps of the low-resolution loss as the training error. During training, we want to minimize the difference between the expensive high-resolution simulations and the simulations produced by the model on a coarse grid. We achieve this by supervised training, where we use the cumulative point error between the predicted conserved component and the true component values as a loss function:
\begin{equation}
Loss =\sum_{i=1}^T \operatorname{MSE}\left(Q_{exact}\left(t_i\right), Q_{pred}\left(t_i\right)\right),
\end{equation}

where MSE denotes mean squared error. Note here $T=32$, $Q_{exact}$ is the true value at each time step and $Q_{pred}$ is the state obtained from NN. The experimental details are shown in section 5.3.

\paragraph{Calculate reconstruction slopes with CNN generated linear coefficients}~\\
In this method, we use CNN to obtain a set of constraint alpha for each conserved component's reconstruction slopes. Note that we can also initialize these coefficients to ensure first-order accuracy. After multiplying the corresponding stencils and reconstruction, we can obtain the boundary values $Q_r$ and $Q_l$. In the time integral, the same approach as the previous one is used to ensure the stability of the solution. However, it is not guaranteed that the interpolated height must be positive. Therefore additional losses should be set during the training process. Another approach is to use Lax-friedlich numerical scheme in (2.11)~\cite{ref5}, but it has a relatively large error compared to other numerical fluxes. In addition, we can add a total energy loss function to increase stability. We introduce the experiments of this approach in section 5.4.
\begin{figure}[H]
\centering
\includegraphics[width=1\textwidth]{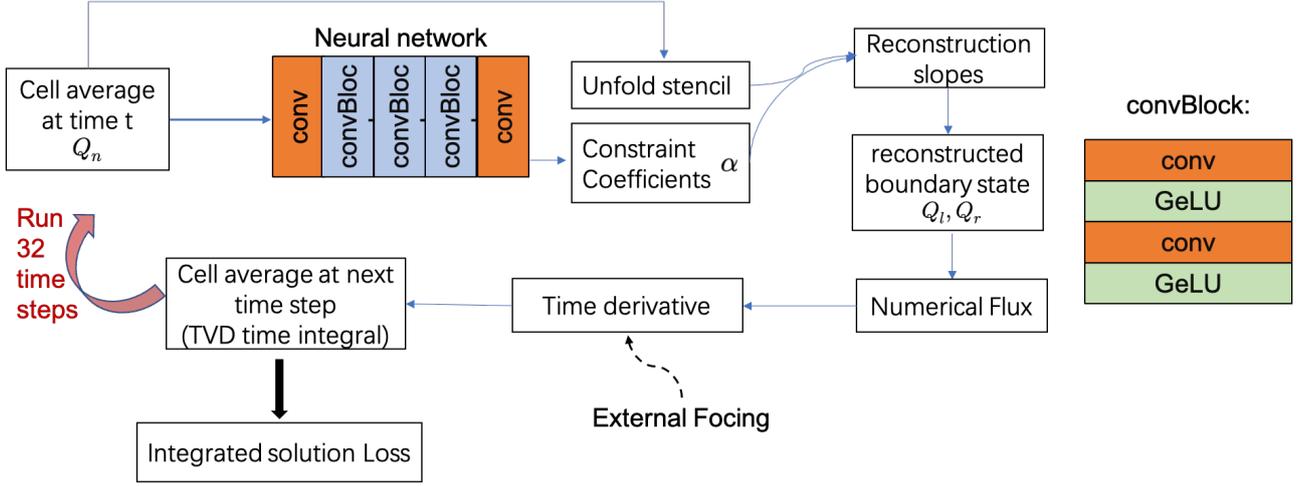}
\caption{\label{fig:nn4}Fourth NN structure: Calculate reconstruction slopes with CNN generated linear coefficients.}
\end{figure}

\section{Classic Solver for SWEs}
In this chapter, we test the classic SWE solvers in 1D, 2D, and spherical for each numerical flux and compare them with Pyclaw's. We use the relative error in height term: $$Error = \frac{\sum_{i=1}^{n}\frac{h_{i}^{high}-h_{i}^{low}}{h_{i}^{high}}}{n} ,$$ where $n$ denote the number of the cells, $h_{i}^{high}$ is the solution of coarsened high resolution in Pyclaw. Furthermore, $h_{i}^{low}$ is the test solution under our solver. In addition, we compare the runtime on CPU and GPU in Pyclaw, torch, and Dace frameworks.
\subsection{One dimensional SWE}
The one-dimensional SWE we introduced in 2.1, where $q=(h, h u)^T$ represents the state vector with the height $h$ and Momentum. The flux vector present:
$$    F(q) = \left(\begin{array}{c}h u \\ h u^2 +\frac{1}{2} g h^2\end{array}\right).$$
\subsubsection{First order solver}
Here we use a dam break problem as the test case in the domain $x \in [0,1]$, and PBC(Periodic Boundary Conditions) is used to be the boundary condition.
$$h=\begin{cases}1&x< 0.5,\\ 0.35& x\geqslant 0.5 .\end{cases}$$
Figure 8 shows the results with the numerical fluxes introduced in 2.3.2 running for 1s. The reference solutions are obtained based on the HLLE sharpclaw solver on a fine grid $2048 \times 2048$ from Pyclaw.
\begin{figure}[H]
\centering
\includegraphics[width=1\textwidth]{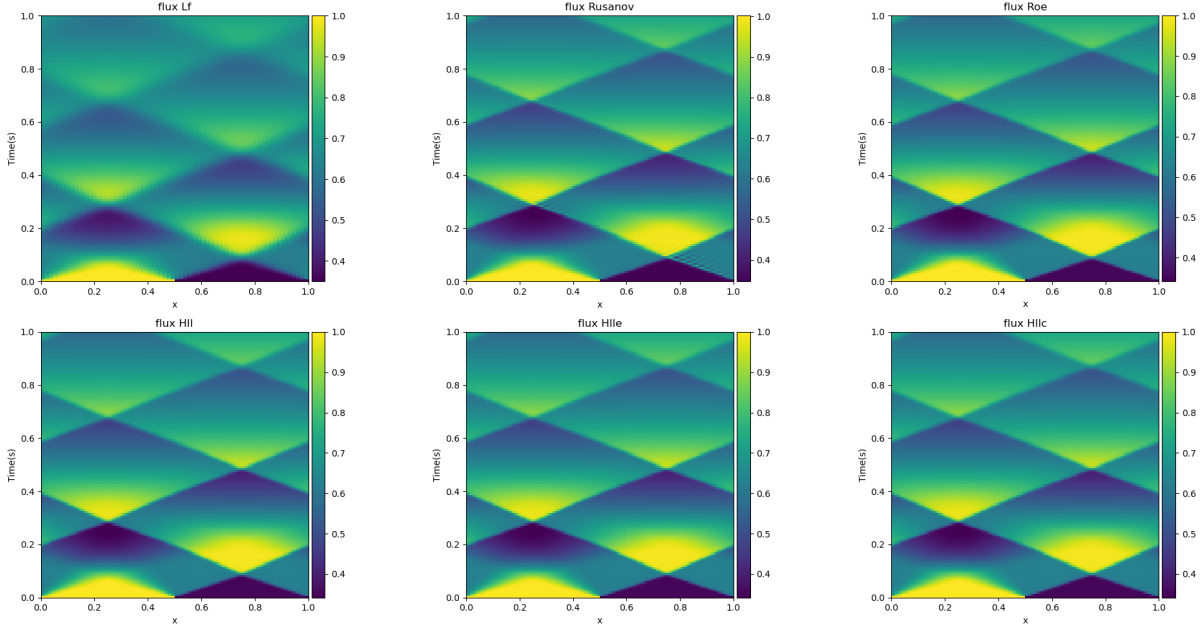}
\caption{\label{fig:SWE1d1order}Solution of the first order one dimensional SWE riemann problems with resolution $128\times 128$ using different numerical flux for the test cases.}
\end{figure}

\begin{table}[H]
    \centering
    \begin{tabular}{|l|cccccc|}
    \hline
    \diagbox{nx}{Error}{Flux} &LF & Rusanov & Roe & HLL & HLLE & HLLC \\
    \hline
    64      &9.15   \%& 2.92  \%& 2.75  \%& 2.81  \%& 2.75  \%&2.78\% \\
    128     &5.15   \%& 1.50  \%& 1.48  \%& 1.51  \%& 1.47  \%&1.49\% \\
    256     &2.67   \%& 0.90  \%& 0.95  \%& 0.96  \%& 0.94  \%&0.95\% \\
    512     &1.36   \%& 0.56  \%& 0.61  \%& 0.62  \%& 0.61  \%&0.61\% \\
    \hline
    \end{tabular}
    \caption{1D SWE $h$ relative error using first order scheme with different numerical fluxes and increased grid.}
    \label{tab:my_label}
\end{table}

Table 1 shows that the relative error of all numerical solutions decreases as the grid increases. Among them, the Lax-Friedlich scheme has the most significant error, and Roe and HLLE perform best at low accuracy. It coincides with Figure 8, where the L-F method gives the most ambiguous results. 

\subsubsection{Second order solver}
We use the same test case as in the previous section and adopt MINMOD slop limiter scheme (2.18) to obtain the numerical solution with high-order accuracy,

\begin{figure}[H]
\centering
\includegraphics[width=1\textwidth]{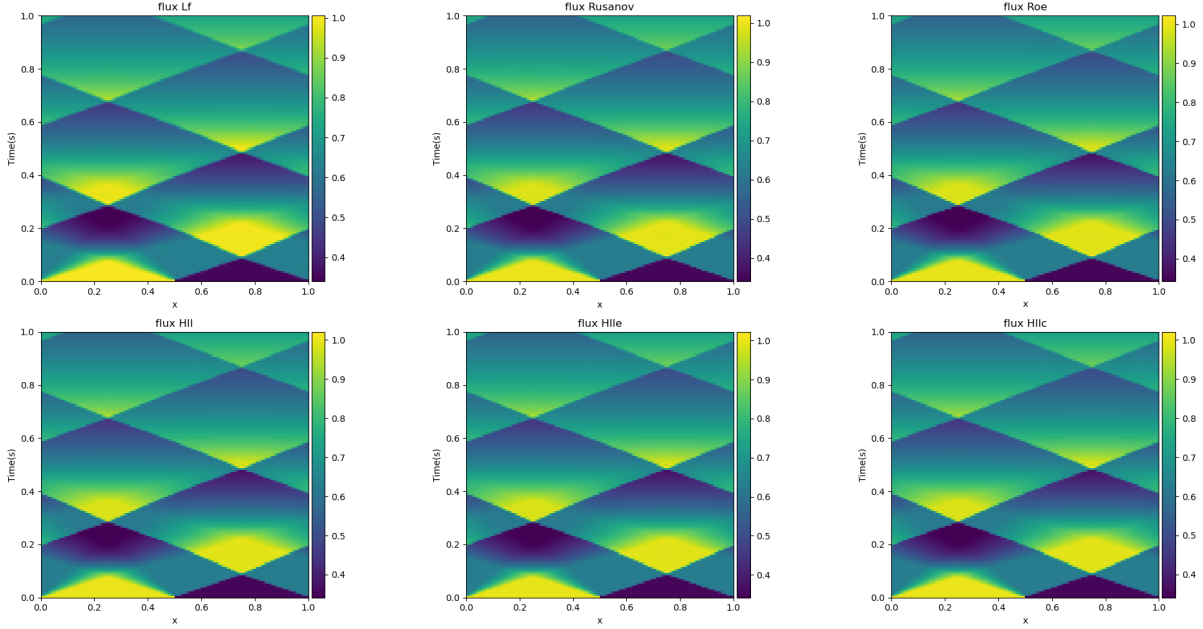}
\caption{\label{fig:SWE1d2order}Solution of the second order one dimensional SWE riemann problems with resolution $128\times 128$ using different numerical flux for the test cases.}
\end{figure}

\begin{table}[H]
    \centering
    \begin{tabular}{|l|cccccc|}
    \hline
    \diagbox{nx}{Error}{Flux} &LF & Rusanov & Roe & HLL & HLLE & HLLC \\
    \hline
    64      &0.99   \%& 0.75  \%& 0.72  \%& 0.73  \%& 0.72  \%&0.73\% \\
    128     &0.46   \%& 0.39  \%& 0.36  \%& 0.37  \%& 0.36  \%&0.36\% \\
    256     &0.34   \%& 0.32  \%& 0.30  \%& 0.30  \%& 0.30  \%&0.30\% \\
    512     &0.27   \%& 0.27  \%& 0.26  \%& 0.26  \%& 0.26  \%&0.26\% \\
    \hline
    \end{tabular}
    \caption{1D SWE $h$ relative error using second order scheme with different numerical fluxes and increased grid.}
    \label{tab:my_label}
\end{table}

Figures 8 and 9 show that the results of the second-order scheme are better than the first-order method. At low resolution, the relative error is even reduced by more than four times. Furthermore, it coincides with the results of the first-order method, Lax-Friedlich still performs the worst, and the relative error is almost the same in the rest of the numerical flux scheme. In the following sections, we will compare the results using the three second-order numerical flux schemes: Rusanov, Roe, and HLLE.

\subsection{SWEs on Planar}
The two-dimensional SWE in Cartesian coordinates $\vec{x} = (x, y)$ can be formulated as:
$$\partial_t q+\nabla \cdot F(q)=S(\vec{x}, q),
$$
where $q=(h, h u, h v)^T$ represents the state vector composed of the height $h$ and two Cartesian velocity components $\vec{u}=(u, v)^T$ all being functions of space and time. The flux matrix has the form
$$
F(q)=\left(\begin{array}{cc}
h u & h v \\
h u^2+\frac{1}{2} g h^2 & h u v \\
h u v & h v^2+\frac{1}{2} g h^2 \\
\end{array}\right).
$$
When the river bottom is not planar, The source term will be $S(\vec{x})$, which acts only on the momentum, which is presented in (2.3),
$$
S(x) = 
\left(\begin{array}{c}
0 \\
-g h \frac{\partial z}{\partial x}(x, y) \\
-g h \frac{\partial z}{\partial y}(x, y)
\end{array}\right).
$$
\subsubsection{Second order solver without source term}
First, we use a Stillwater problem as the test case in the domain $x \times y\in[0,1]\times[0,1]$. For the height term we use a two-dimensional Gaussian distribution $\exp(-((x-0.5)^2+(y-0.5)^2)/\sigma^2) + 0.5, \sigma = 0.5$ as the initial condition, PBC(Periodic Boundary Conditions) is considered as the boundary condition. Figure 10 shows the height obtained on a grid with $400\times400$ grid cells with the numerical fluxes introduced in 2.3.2 running for 1.5$s$. The reference solutions are obtained based on the HLLE sharpclaw solver on a fine grid $1000 \times 1000$ from Pyclaw.

\begin{figure}[H]
\centering
\includegraphics[width=1\textwidth]{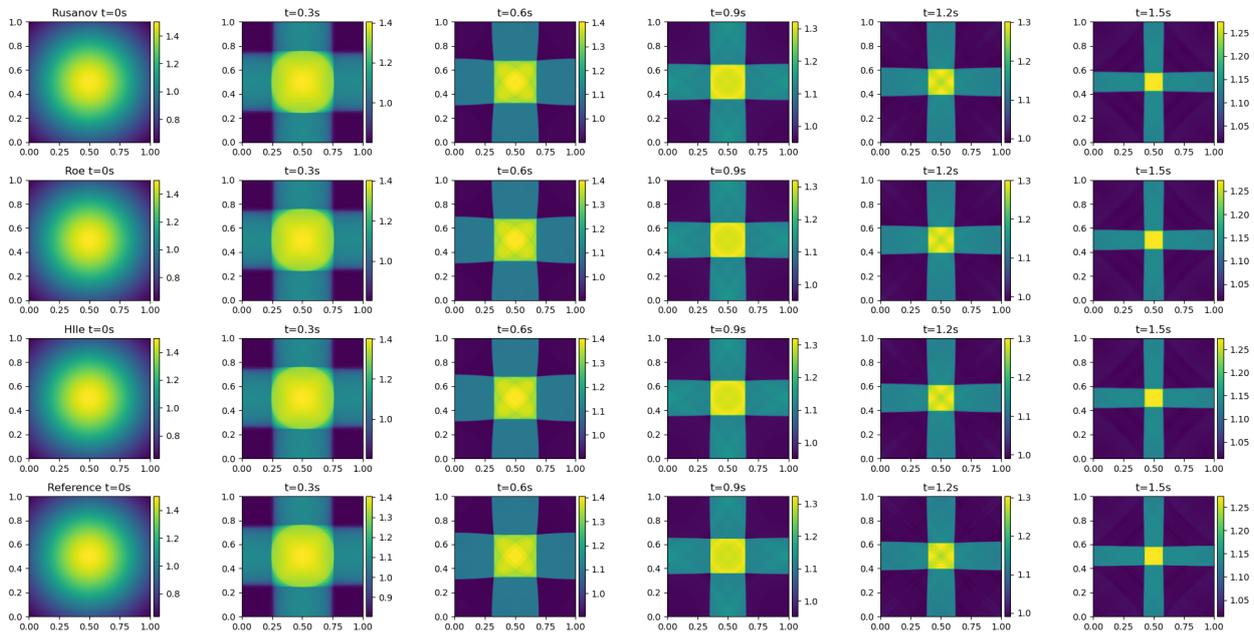}
\caption{\label{fig:SWE_planar}Solution of the 2D SWE with resolution $400\times 400$ using different numerical flux for the test cases.}
\end{figure}

\begin{table}[H]
    \centering
    \begin{tabular}{|l|ccccc|}
    \hline
    \diagbox{nx}{Error}{Flux} & Rusanov & Roe & HLL & HLLE & HLLC \\
    \hline
    $100\times100$      & 0.44  \%& 0.43  \%& 0.43  \%& 0.44  \%&0.43\% \\
    $200\times200$      & 0.19  \%& 0.19  \%& 0.19  \%& 0.19  \%&0.19\% \\
    $400\times400$      & 0.08  \%& 0.07  \%& 0.07  \%& 0.07  \%&0.07\% \\
    $800\times800$      & 0.05  \%& 0.05  \%& 0.05  \%& 0.05  \%&0.05\% \\
    $1000\times1000$     & 0.05  \%& 0.03  \%& 0.04  \%& 0.03  \%&0.03\% \\

    \hline
    \end{tabular}
    \caption{2D SWE $h$ relative error using second order scheme with different numerical fluxes and increased grid.}
    \label{tab:my_label}
\end{table}

\subsubsection{Second order solver with the source term}
We generate the water bed $z(x, y)$ using two-dimensional Perlin noise, as shown in Figure 11. Furthermore, we use the same initial condition as the test case. Here we use Roe as the numerical flux and wall Boundary Condition. The reference solution is from the bathymetry solver on $1000\times1000$  grid from Pyclaw.
\begin{figure}[H]
    \centering
    \subfigure[2D water bed]{
        \includegraphics[width=2.5in]{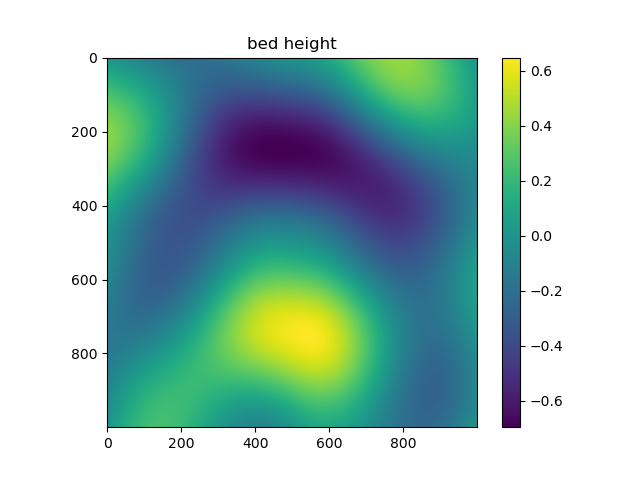}
    }
    \subfigure[3D water bed and water surface]{
	\includegraphics[width=2.1in]{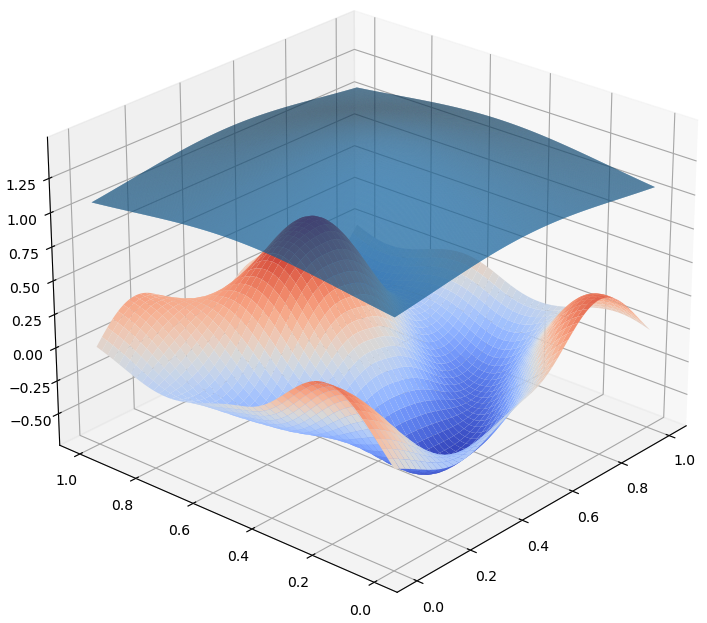}
    }
    \caption{Water bed generated by 2D Perlin noise.}
    \label{fig.1}
\end{figure}

\begin{figure}[H]
\centering
\includegraphics[width=1\textwidth]{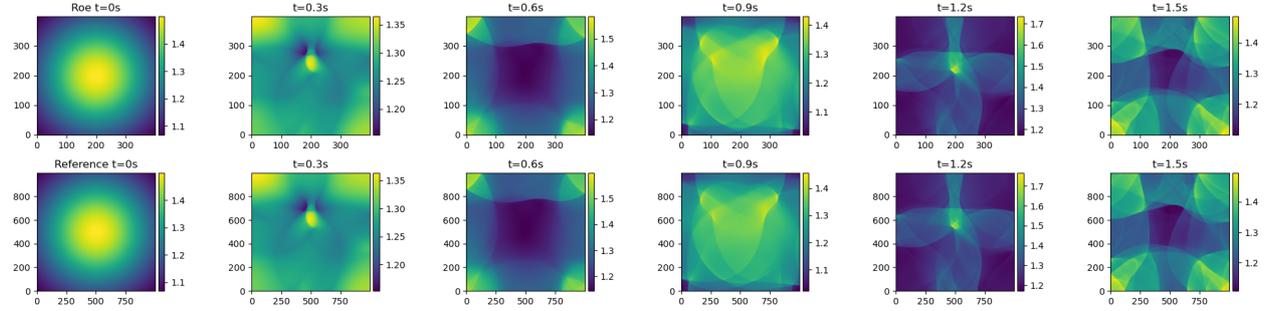}
\caption{\label{fig:SWE_planar_height}Solution of the 2D SWE with water bed on resolution $400\times 400$.}
\end{figure}

The resulting Figure 12 shows that our solver is consistent with the one in Pyclaw.

\subsection{SWEs on Sphere}
The shallow water equations are often solved on a sphere in the global atmosphere or ocean models, for example. The equations are formulated in three-dimensional Cartesian coordinates $\vec{x}=(x, y, z)$ now, where $q=(h, h u, h v, h w)^T$ represents the state vector composed of the height $h$ and three Cartesian velocity components $\vec{u}=(u, v, w)^T$. The flux matrix has the form:
$$
F(q)=\left(\begin{array}{ccc}
h u & h v & h w \\
h u^2+\frac{1}{2} g h^2 & h u v & h u w \\
h u v & h v^2+\frac{1}{2} g h^2 & h v w \\
h u w & h w & h w^2+\frac{1}{2} g h^2
\end{array}\right).
$$
In order to obtain a conservative update, we should also subtract the correction term in each grid cell average value of $Q^{n}$ during each time step. If we let all the normal vectors outward pointing, then we have:
\begin{equation}
Q^{cor}=-F(Q^{n})\cdot\sum_{i=1}^{N}h_{i}\vec{n}_{i}.
\end{equation}
The source term $S(x, q)$, which acts only on the momentum equations, has the form:
\begin{equation}
S(x, q)=-\frac{2 \omega}{a} z(x \times h u)+(x \cdot(\nabla \cdot \tilde{f})) x.
\end{equation}
The first source term on the right-hand side of $(2.1)$ is the Coriolis force due to the earth's rotation. Here $\omega$ and $a$ are the angular speed and the earth's radius, respectively. The second source term in $(2.1)$ is a forcing term that guaranties that the fluid velocity remains on the surface of the sphere, i.e., perpendicular to the position vector $\vec{x} = (x, y, z)$ on the sphere. Here $\tilde{f}$ consists of the part of the flux matrix associated with the momentum equations.
The Cartesian form has the advantage that it is independent of the coordinate transformation being used. The approximation of the three-dimensional equations is restricted to the surface of the sphere, i.e., we need to update $Q$ in grid cells on the sphere. Our discretization follows the general outline from section 2.3.4. The details are as follows,

\begin{enumerate}
    \item Compute the cell edges length $h_{i}$ and cell area $\Delta c$, which here are the distance and area from physical domain. Great-circle distance and area are considered.
    \item Define the direction normal $\vec{n}$ and tangential $\vec{t}$ on each side of an interface into the interface in the tangent plane; where $\vec{t}$ can be defined using the edge vetor and $\vec{n}$ can be calculated by the cross product of $\vec{t}$ and centripetal vector of $\vec{t}$, now $\vec{n}, \vec{t} \in \vec{\mathbb{R}}^3$. 
    \item Rotate the velocity components of the grid cell values using rotate matrix $R$ from (2.24).
    \item Solve the one-dimensional Riemann problem and the momentum components of the flux $\bar{F}^{n\bullet}$ are rotated back to Cartesian coordinates $\bar{F}^{n}$ using $R^{T}$.
    \item Repeat 1-4 step to obtain all cell's interface flux.
    \item Calculated the correction term (3.1) and Coriolis force (3.2) using a higher order Runge-Kutta method.
    \item Update all the cells' state, then for each time step the momentum components are projected to the tangent plane at each cell center.
\end{enumerate}

In the next, we use Rossy-Haurwitz as the test problem. In our following example, we consider approximations of a wave number four Rossby-Haurwitz pattern, a standard test problem for the shallow water equations on a rotating sphere. We refer to Williamson et al.~\cite{ref25} for a detailed description of the initial conditions. Haurwitz shows that the flow pattern moves from west to east with constant angular velocity and no change in shape. Although the Rossby-Haurwitz wave is not an analytic solution of the shallow water equation, the wave structure is rotated around the Z-axis for a long time. For our simulations, we have set the characteristic time scale to one day and the characteristic length scale to the earth's radius.

Figure 13 shows the height obtained on a grid with $1000\times500$ grid cells at different times. Here we use the second-order solver of the Roe flux scheme with (MINMOD) flux limiter.
\begin{figure}[H]
\centering
\includegraphics[width=0.6\textwidth]{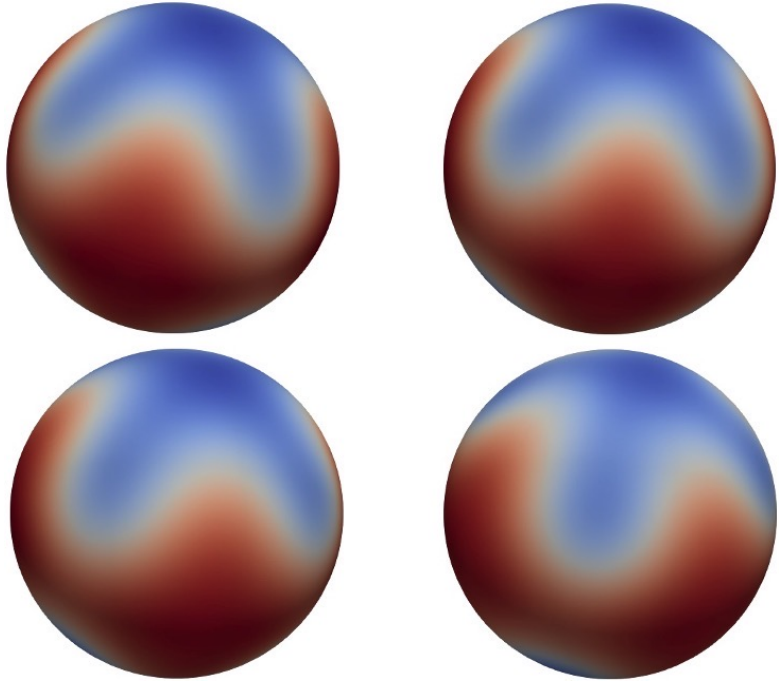}
\caption{\label{fig:computational time}Shallow water equations on sphere with Rossby-Haurwitz wave as initial data. The fluid height $h$ are shown. The top two figures (times t=0 and 1 days) are viewed from above. The bottom two figures (times t=2 days and 4 days) are viewed from below.}
\end{figure}

\begin{table}[H]
    \centering
    \begin{tabular}{|l|ccccc|}
    \hline
    \diagbox{nx}{Error}{Flux} & Rusanov & Roe & HLL & HLLE & HLLC \\
    \hline
    $100\times50$       & 0.42  \%& 0.37  \%& 0.42  \%& 0.42  \%&0.37\% \\
    $200\times100$      & 0.15  \%& 0.14  \%& 0.15  \%& 0.15  \%&0.14\% \\
    $400\times200$      & 0.07  \%& 0.07  \%& 0.08  \%& 0.08  \%&0.07\% \\
    \hline
    \end{tabular}
    \caption{Rossby-Haurwitz wave using three grids with $100\times 50$, $200 \times 100$ and $400 \times 200$ grid cells and different numerical flux scheme. The table shows the $h$ relative error at 1 day.}
    \label{tab:my_label}
\end{table}

\subsection{Computational effort on different framework}
We implemented the above classic SWE solver on Dace(Data-Centric Parallel Programming)~\cite{ref26} and Torch frameworks. This section compares their speed performance with the solver from Pyclaw. Note that Pyclaw does not support calculations under the GPU framework, which is a disadvantage when dealing with a complicated problem. Here we uniformly use the second-order accuracy roe Riemann solver. We also give the test on the different devices where CPU we use AMD EPYC 7742 @ 2.25GHz and NV RTX3090 for the GPU test device. \\
The result of the planar SWE solver is shown in Figure 14. Under the DACE framework, we need 100s pre-compile the solver code. After the pre-compilation, we can enter an arbitrary period and resolution for the computing.
\begin{figure}[H]
\centering
\includegraphics[width=0.7\textwidth]{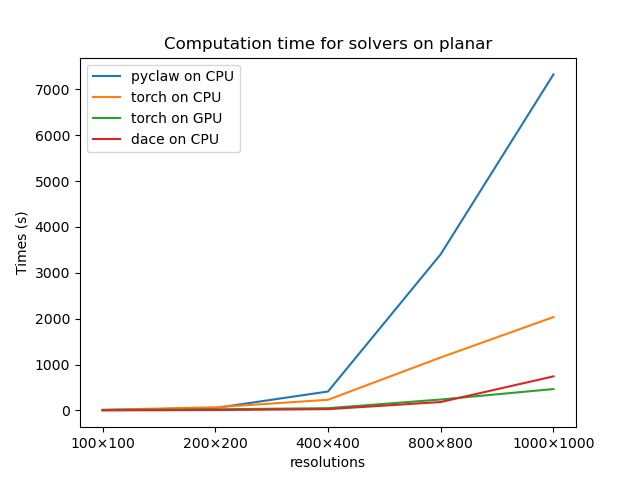}
\caption{\label{fig:computational time}Compare the performance of the 2D SWE solvers on different framework. The resolution is from $100 \times 100$ to $1000\times1000$. }
\end{figure}
We can see that our solver performs much better than Pyclaw in either framework. Dace outperforms Torch when running under the CPU, and although the code takes some time to pre-compile when running under Dace, this can be a huge advantage when dealing with high-precision multi-timesteps. Furthermore, we use the same device and methods to compare the computational efficiency of the solver on the sphere under different frameworks. The SWE solver on the sphere is much more complicated than the planar solver, so we need more time for pre-compilation under the dace framework, which takes about 400s.

\begin{figure}[H]
\centering
\includegraphics[width=0.7\textwidth]{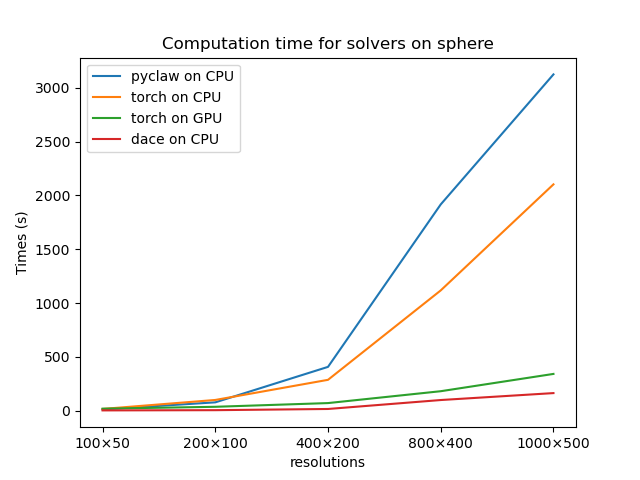}
\caption{\label{fig:computational time on sphere}Compare the performance of the SWE spherical solvers using the Rossy-Haurwitz for 1.5s. The resolution is from $100\times50$ to $1000\times500$.}
\end{figure}

From Figure 15, we can see that all of our solvers outperform Pyclaw's spherical solver in terms of speed under different devices. The highest resolution of Pyclaw's spherical solver is $1000\times500$, while our solver can be of arbitrary accuracy. It is essential when high-precision atmospheric data needs to be modeled. The solver on the sphere is much more complex than the planar. Surprisingly we found that the spherical solver ran even better on CPU on the Dace framework than on torch on GPU. When dealing with complicated problems, efficiency could be improved using the Dace framework.

\section{Classic Solver for Euler equations}

\subsection{One dimensional Euler equation}
The one-dimensional Euler equation presented in 2.2 where $q=(\rho, \rho u, E)^T$ represents the state vector with the density $\rho$ and Momentum. The flux vector present:
$$
F(q)=\left(\begin{array}{c}
\rho u \\
\rho u^2+p \\
u(E+p)
\end{array}\right)=\left(\begin{array}{c}
q_2 \\
q_2^2 / q_1+p\left(q_1, q_2, q_3\right) \\
\left(q_2 / q_1\right)\left(q_3+p\left(q_1, q_2, q_3\right)\right)
\end{array}\right),
$$
where $p\left(q_1, q_2, q_3\right)=(\gamma-1)\left(q_3-\frac{1}{2} q_2^2 / q_1\right)$.

\subsubsection{First order solver}
The following test initial condition with $\gamma=7/5$ is used,
$$
q=\begin{cases}(1, 0, \frac{1}{\gamma-1})^T&x< 0.5,\\ (0.125, 0, \frac{0.1}{\gamma-1})^T& x\geqslant 0.5. \end{cases}
$$
Figure 16 shows the results with the numerical fluxes introduced in 2.3.2 running for 1s. The reference solutions are based on a fine grid $1024 \times 1024$ from the Pyclaw HLLE sharpclaw solver.

\begin{figure}[H]
\centering
\includegraphics[width=1\textwidth]{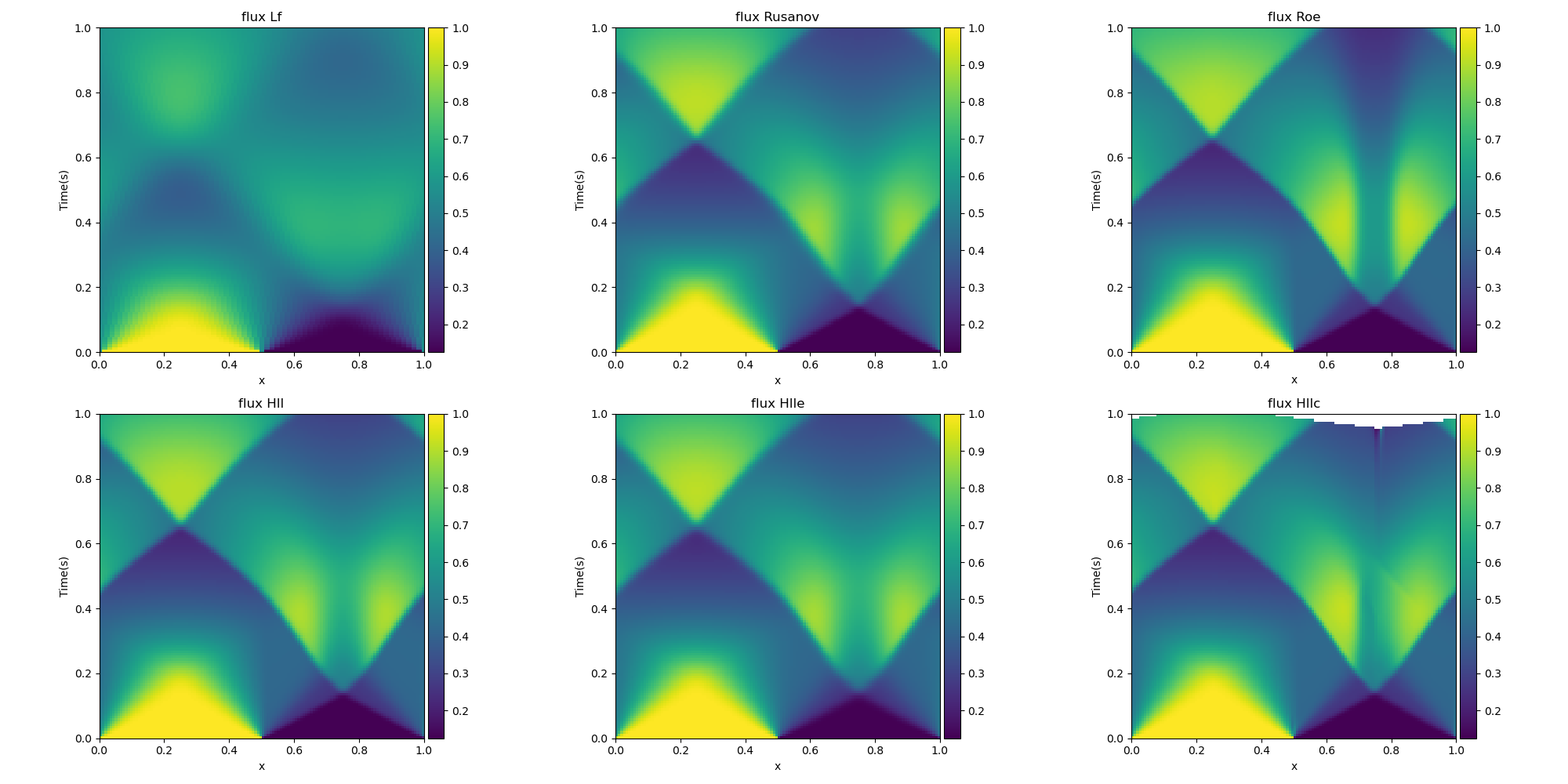}
\caption{\label{fig:euler1d}Solution of the first order one dimensional Euler equation with resolution $128\times 128$ using different numerical flux for the test case.}
\end{figure}

From figure 16, we can see, same as the SWE solver in chapter 3, the L-F method obtains the most ambiguous solution. Note that when at the end, the HLLC scheme solver generates an unstable solution.

\begin{table}[H]
    \centering
    \begin{tabular}{|l|cccccc|}
    \hline
    \diagbox{nx}{Error}{Flux} &LF & Rusanov & Roe & HLL & HLLE & HLLC \\
    \hline
    64      &52.52  \%& 22.41 \%& 10.96 \%& 19.91  \%& 23.09  \%&17.52 \%\\
    128     &42.58  \%& 14.82 \%& 6.54  \%& 13.69  \%& 15.65  \%&Nan \\
    256     &29.68  \%& 9.91  \%& 4.25  \%& 9.64   \%& 10.49  \%&Nan \\
    512     &19.33   \%& 7.05  \%& 2.66  \%& 6.72   \%& 7.50   \%&Nan \\
    \hline
    \end{tabular}
    \caption{1D Euler equation $\rho$ relative error using first order scheme with different numerical fluxes and increased grid. }
    \label{tab:my_label}
\end{table}

According to table 5, the relative error of all numerical solutions decreases when the grid increases. Furthermore, the Lax-Friedlich scheme has the most significant error, and the Roe scheme performs best at all resolutions. Since the HLLC scheme produces an unstable solution at the last moment, the error cannot be calculated.

\subsubsection{Second order solver}
We take the same second-order approach in the SWE solver to achieve a high-order Euler equation solver.

\begin{figure}[H]
\centering
\includegraphics[width=1\textwidth]{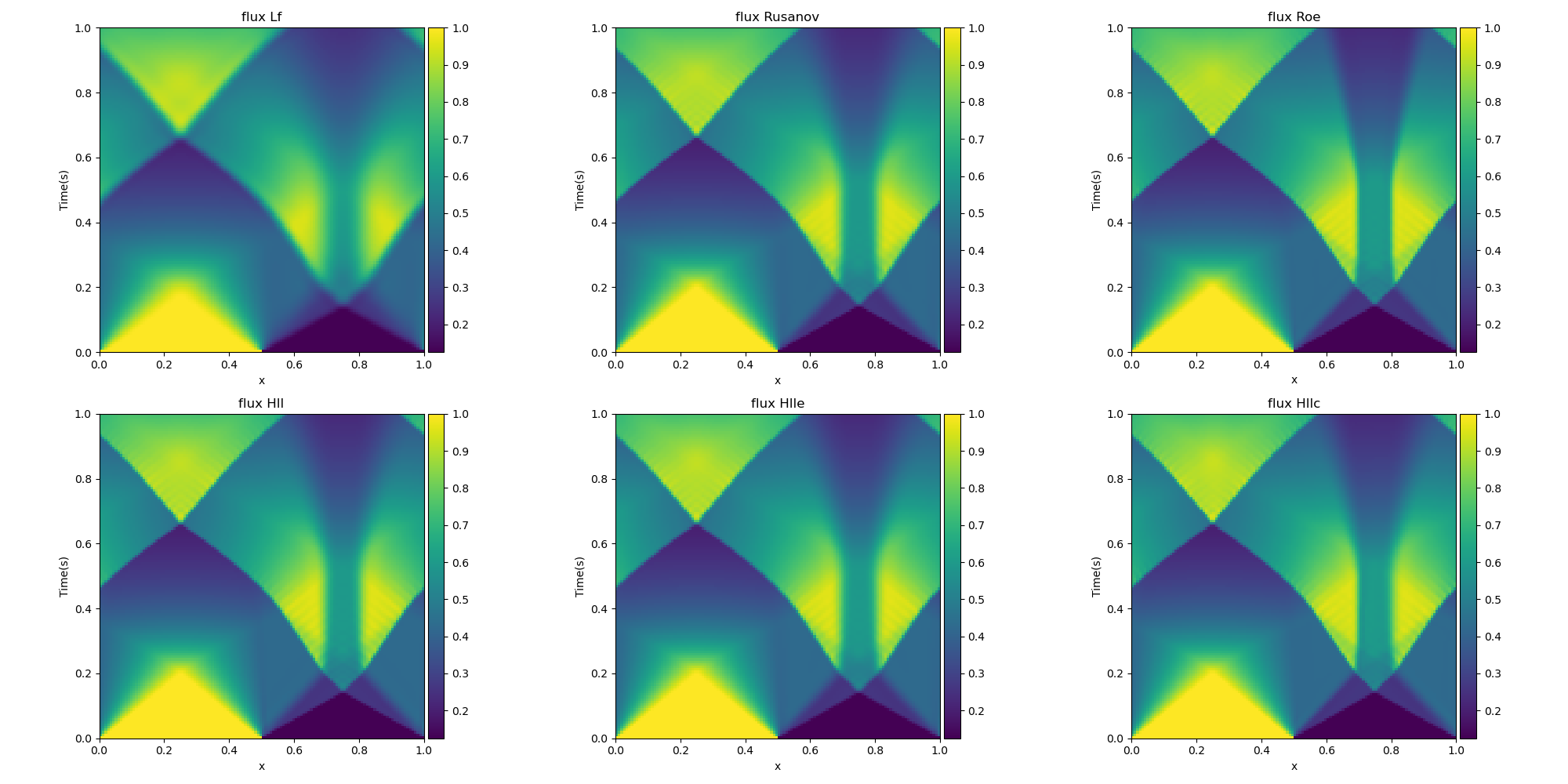}
\caption{\label{fig:euler1d2order}Solution of the second order one dimensional Euler equation with resolution $128\times 128$ using different numerical flux for the test cases.}
\end{figure}

\begin{table}[H]
    \centering
    \begin{tabular}{|l|cccccc|}
    \hline
    \diagbox{nx}{Error}{Flux} &LF & Rusanov & Roe & HLL & HLLE & HLLC \\
    \hline
    64      &7.65   \%& 4.93  \%& 2.63  \%& 4.89  \%& 5.14  \%&4.36\% \\
    128     &4.88   \%& 3.15  \%& 1.76  \%& 3.10  \%& 3.32  \%&2.90\% \\
    256     &2.19   \%& 1.63  \%& 0.68  \%& 1.60  \%& 1.73  \%&1.44\% \\
    512     &1.53   \%& 0.70  \%& 0.37  \%& 0.68  \%& 0.73  \%&0.57\% \\
    \hline
    \end{tabular}
    \caption{1D Euler equation $\rho$ relative error using second order scheme  with different numerical fluxes and increased grid.}
    \label{tab:my_label}
\end{table}

Figures 16 and 17 show that the results of the second-order scheme are much more precise than those of the first order. Moreover, Tables 5 and 6 find that the relative error is even reduced by eight times at second-order accuracy. Moreover, the second-order HLLC scheme produces no anomalous solution at the end. As with the SWE solver, the roe is still the best solution. 

\subsection{Euler equation on Planar}
The two-dimensional Euler equations in Cartesian coordinates $\vec{x} = (x, y)$ can have the conservation vector $q=(\rho, \rho u, \rho v ,E)^T$ , which represent the density $\rho$, energy $E$ and two Cartesian velocity components $\vec{u}=(u, v)^T$ all being functions of space and time. The flux matrix has the form:
$$
F(q)=\left(\begin{array}{cc}
\rho u & \rho v \\
\rho u^2+p & rho u v \\
\rho u v & \rho v^2+p \\
u(E+p)&v(E+p)\\
\end{array}\right)
$$
The initial condition we used is one of the 2D Riemann problems from the paper of Liska \& Wendroff (2003)~\cite{ref27}, which is also the example in Pyclaw and Wall boundary condition is set. Figure 18 shows the density obtained on a grid with $400\times400$ grid cells with the Rusanov, HLL, and HLLE numerical fluxes running for 0.3$s$. We use the same slop flux limiter in the second-order solver as SWEs. Moreover, the reference solutions are obtained based on the HLLE sharpclaw solver on a fine grid $1000 \times 1000$ from Pyclaw.

\begin{figure}[H]
\centering
\includegraphics[width=1\textwidth]{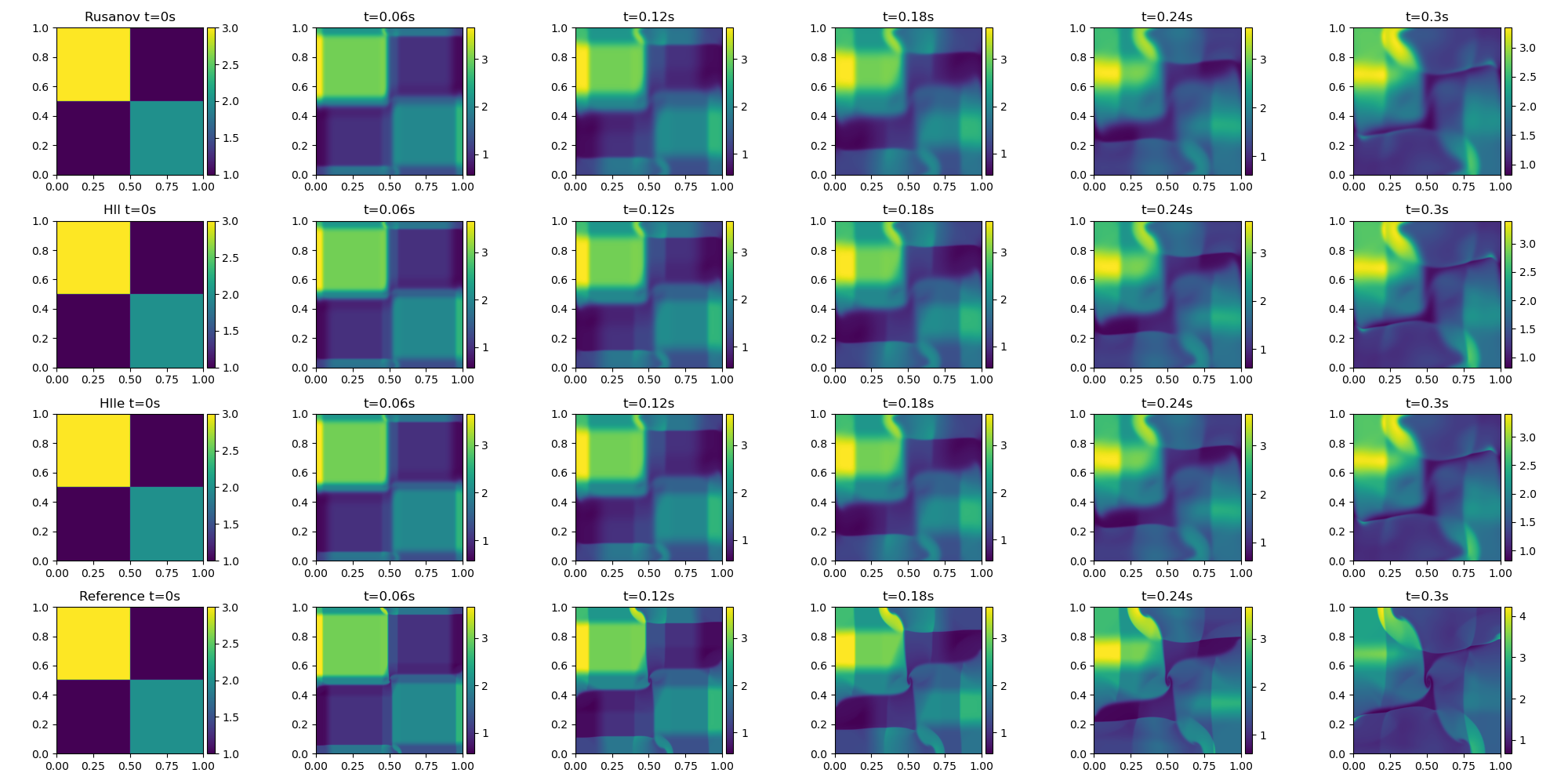}
\caption{\label{fig:euler_planar1oder}Solution of the first order 2D Euler riemann problems with resolution $400\times 400$ using different numerical flux for the test cases.}
\end{figure}

\begin{table}[H]
    \centering
    \begin{tabular}{|l|cccc|}
    \hline
    \diagbox{nx}{Error}{Flux} & Rusanov & Roe & HLL & HLLE \\
    \hline
    $100\times100$      & 10.99  \%& 6.35  \%& 10.46  \% & 10.44 \%\\
    $200\times200$      & 8.72   \%& 4.29  \%& 8.05  \% & 9.05 \%\\
    $400\times400$      & 6.43   \%& 2.58  \%& 5.82  \% & 5.85 \%\\
    \hline
    \end{tabular}
    \caption{2D Euler equation $\rho$ relative error using first order scheme with different numerical fluxes and increased grid.}
    \label{tab:my_label}
\end{table}

\begin{figure}[H]
\centering
\includegraphics[width=1\textwidth]{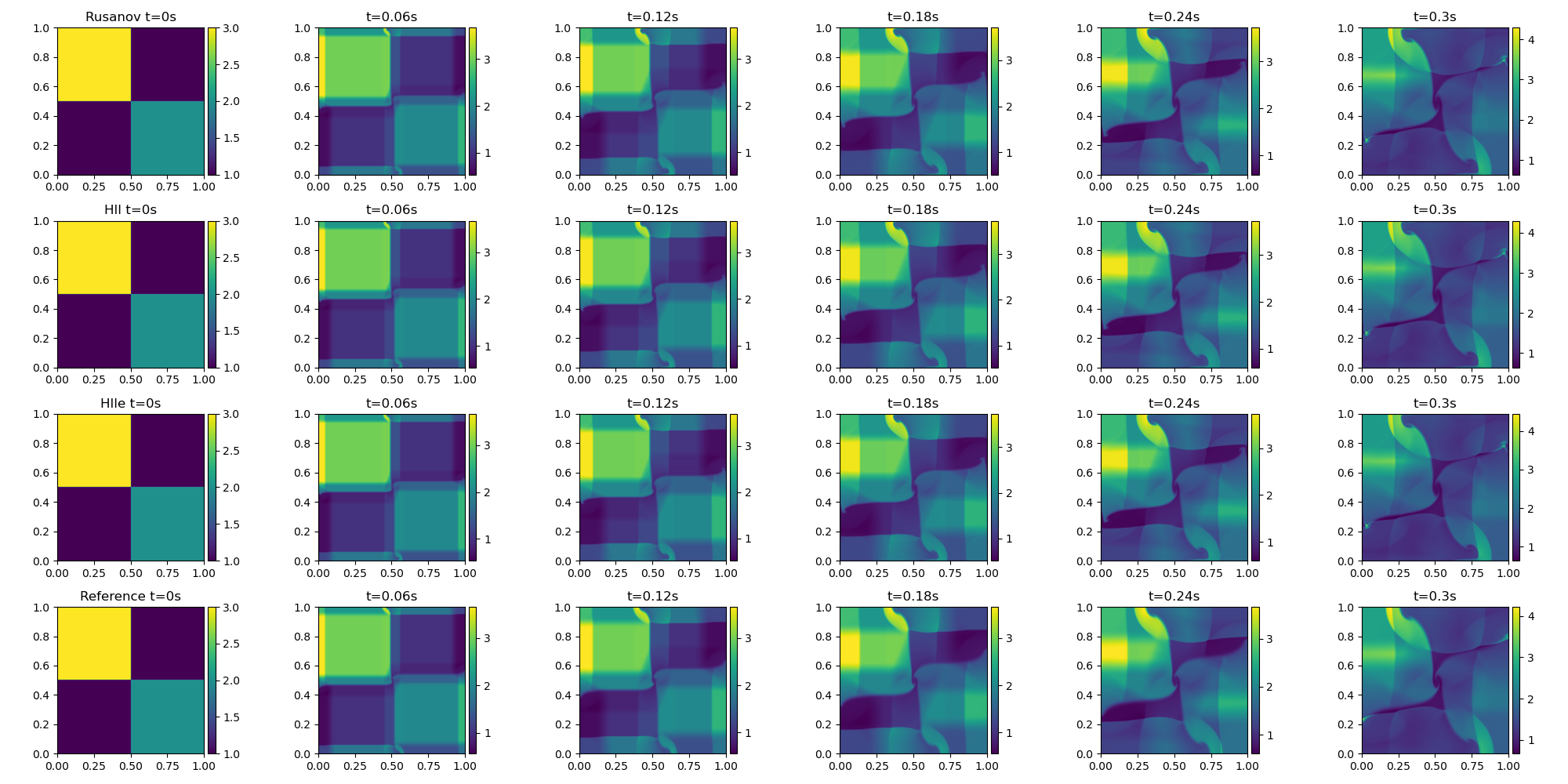}
\caption{\label{fig:euler_planar2order}Solution of the second order 2D Euler riemann problems with resolution $400\times 400$ using different numerical flux for the test cases.}
\end{figure}

\begin{table}[H]
    \centering
    \begin{tabular}{|l|cccc|}
    \hline
    \diagbox{nx}{Error}{Flux} & Rusanov & Roe & HLL & HLLE \\
    \hline
    $100\times100$      & 4.89  \%& 4.60  \%& 4.63  \% & 4.65 \%\\
    $200\times200$      & 2.76  \%& 2.71  \%& 2.76  \% & 2.79 \%\\
    $400\times400$      & 2.08  \%& 2.04  \%& 2.14  \% & 2.16 \%\\
    \hline
    \end{tabular}
    \caption{2D Euler equation $\rho$ relative error using second order scheme with different numerical fluxes and increased grid.}
    \label{tab:my_label}
\end{table}

Comparing Fig. 18 with Fig. 19, we can find that the solution of the second-order solver is significantly sharper than the first-order one, consistent with Tables 7 and 8. Consistent with the SWE solver in 3.4, the computation time of our solver is much faster than that of the plaw solver.

\section{Embedded Data-Driven Solver for SWEs}
 This chapter uses four data-driven methods presented in 2.52 and 2.53 to implement the ML-based SWE solver.
 Furthermore, the total energy and potential enstrophy are used to test the energy dissipation of the classic solver and the data-driven solver~\cite{ref33}, which can be expressed as follows: 

\begin{equation}
\begin{array}{c}
potenial = 0.5*gh^{2}\\
kinetic = 0.5 * (u^2+v^2 + w^2)*h\\
energy = potential + kinetic
\end{array}
\end{equation}

\subsection{Calculate boundary Flux directly from CNN}
\paragraph{SWE on shpere}~\\
We generated a training set of 300 different initial conditions with a resolution of $1000\times500$ using spherical Perlin noise~\cite{ref29}. The order of magnitude of the conserved variables is consistent with the Rossby-Haurwitz wave. Each initial case is then run in Pyclaw for 0.5 days. Finally, we obtained a total number of 200,000 training sets where each state is stable and continuous. An example is shown in Figure 20.
\begin{figure}[H]
\centering
\includegraphics[width=0.9\textwidth]{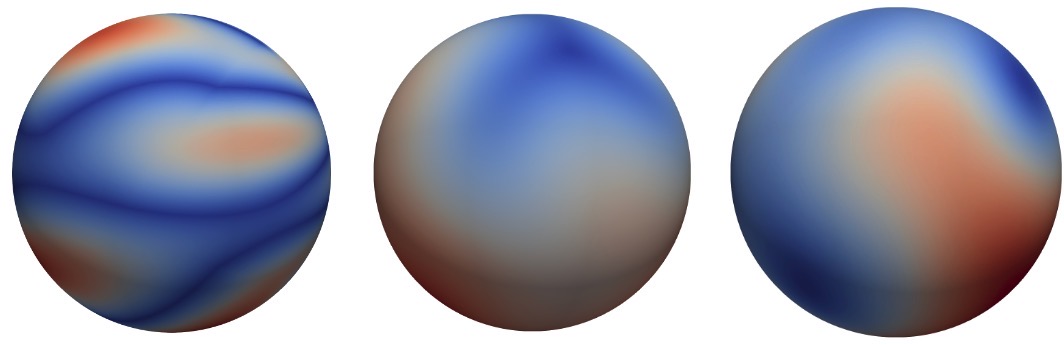}
\caption{\label{fig:trainset_sphere}Spherical training set generated by Perlin noise on sphere.}
\end{figure}
Then we use the roe scheme to compute the flux of the cell after coarsening, as shown in Figure 25. "original" and "center" indicate the flux of each cell in high and coarsened low resolution, respectively.
\begin{figure}[H]
\centering
\includegraphics[width=0.7\textwidth]{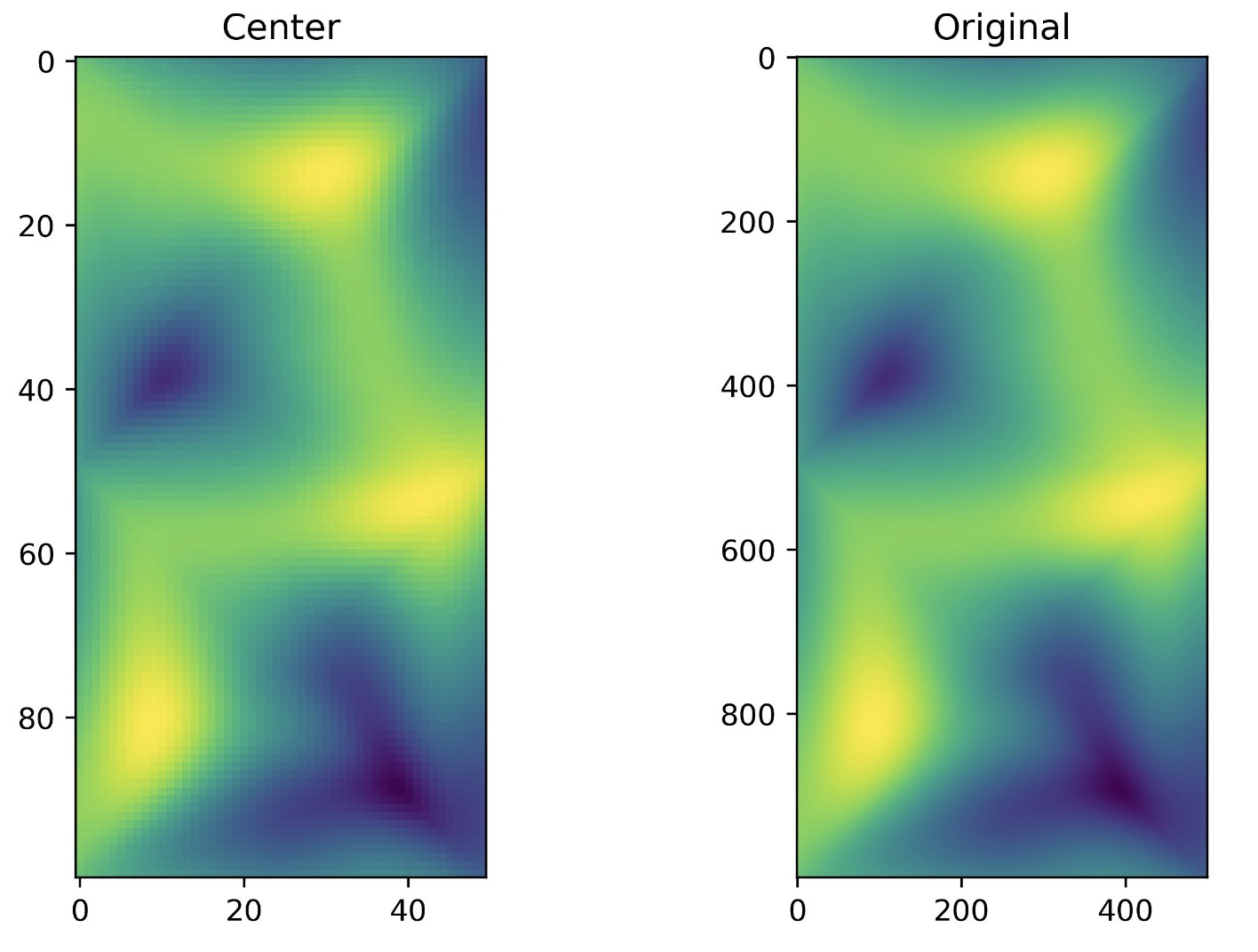}
\caption{\label{fig:trainset_sphere3}Interpolate cell flux from high to low resolution mesh grid on computational grid.}
\end{figure}
After 200 epochs of training, we embed the NN into our classic solver,
\begin{figure}[H]
\centering
\includegraphics[width=1.0\textwidth]{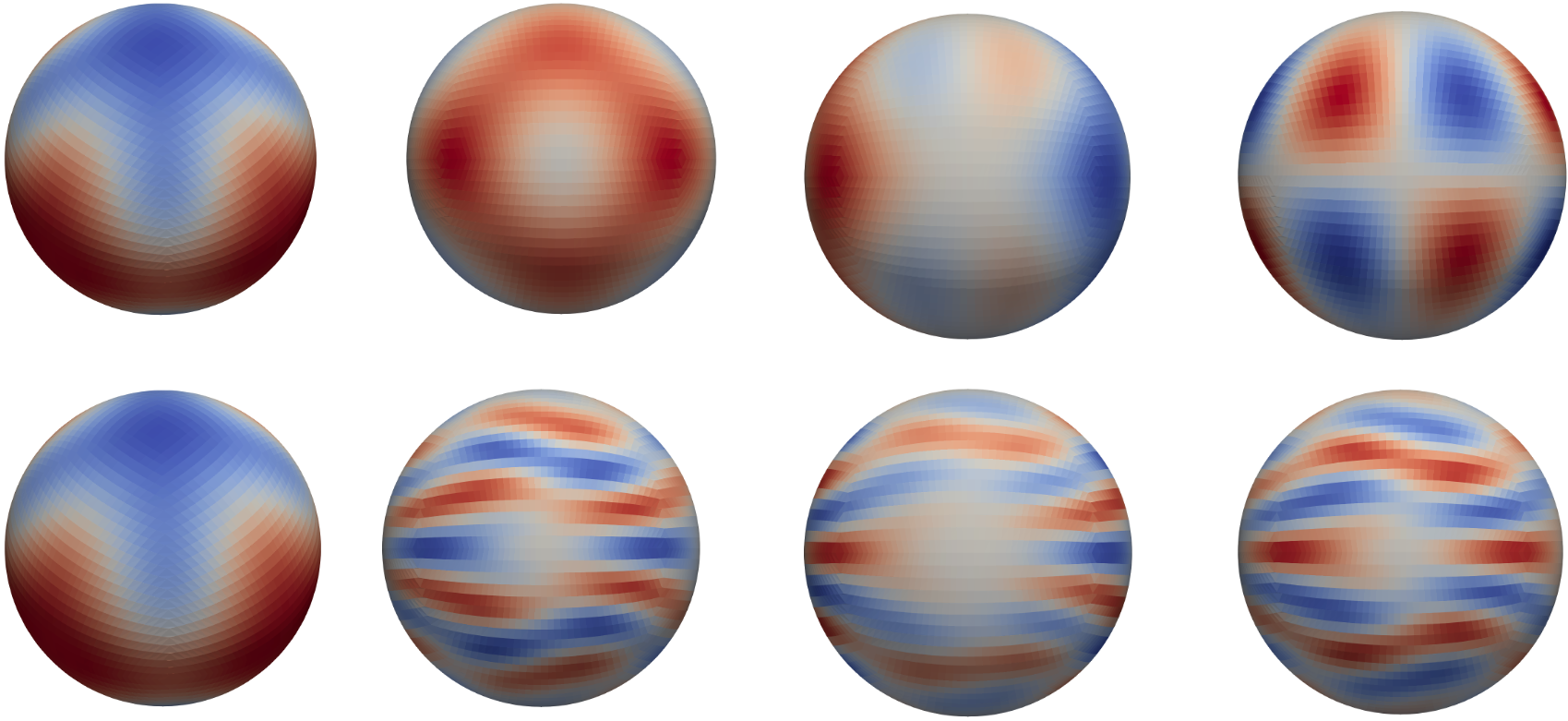}
\caption{\label{fig:test0-sphere}The upper four pictures are conservation vector $(h, hu, hv, hw)^T$ of initial value. The following four pictures are the results using first NN approach running for one day.}
\end{figure}

As we can see in Fig. 22, the solver runs for one day, the height $h$ remains unchanged, and the results for the remaining momentum components $(hu, hv, hw)^T$ are entirely wrong.

\begin{figure}[H]
\centering
\includegraphics[width=1.0\textwidth]{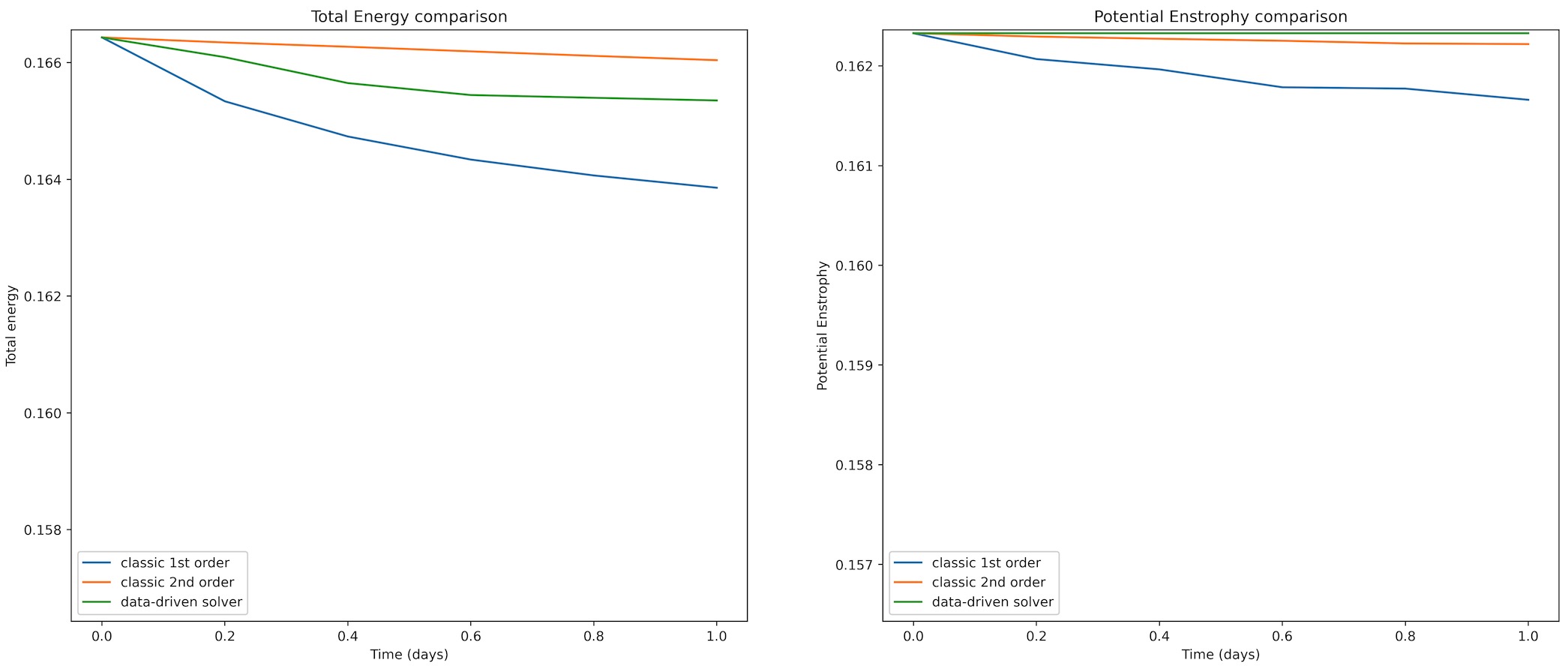}
\caption{\label{fig:test0-sphere}Evolution of the conservation of total energy (left) and potential enstrophy (right) on $100\times50$ mesh grid for Rossby–Haurwitz wave test using first NN approach.}
\end{figure}

After the energy test, we can see that the total energy loss is between the first-order accuracy and the second-order, while the potential enstrophy is unchanged. This NN structure is not feasible, even though the energy conservation is well-behaved.

\subsection{Calculate boundary states with CNN generated linear coefficients}
\paragraph{SWE on shpere}~\\
Here we use the same training set as in the previous section 5.1. Then we tried two interpolation methods (Scipy~\cite{ref31} and Xesmf~\cite{ref32}) to obtain the four boundary values after coarsening the mesh. Figure 24 represents the result of coarsening our grid of $1000\times500$ to $100\times500$ by a factor of ten. The "center" represents the center value in the low-resolution grid, "left," "up," "right," and "bottom" represent the four boundary values, and "original" indicates the center value in the high-resolution grid.

\begin{figure}[H]
\centering
\includegraphics[width=1\textwidth]{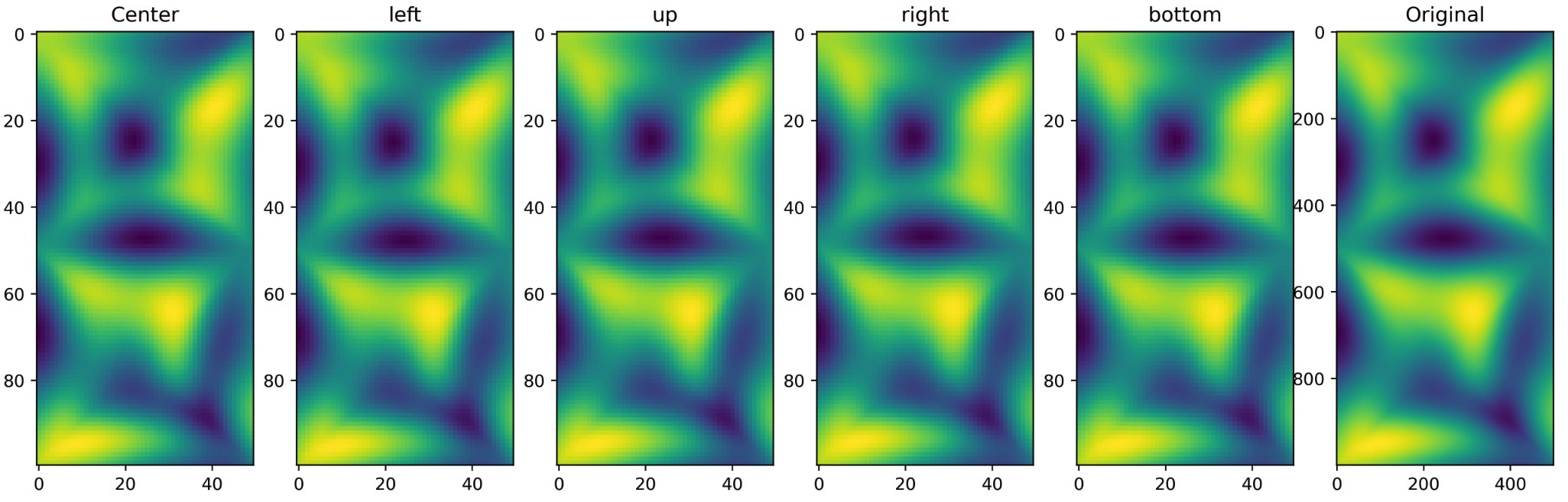}
\caption{\label{fig:trainset_sphere2}Interpolate center values from high to low resolution mesh grid on computational grid.}
\end{figure}
After this, we take the center values of the coarsened low-resolution mesh as input values and the four boundaries as output values. After 200 epochs of training, we embed the NN into our classic solver.

\begin{figure}[H]
\centering
\includegraphics[width=0.8\textwidth]{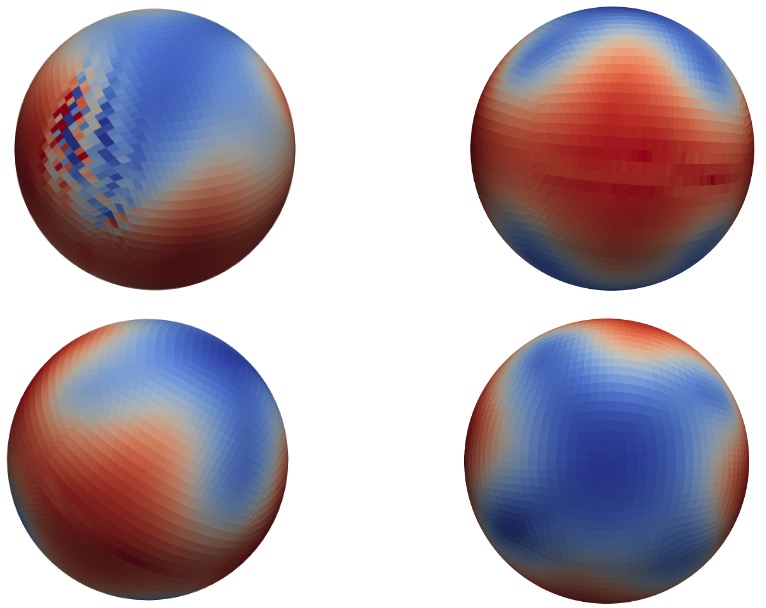}
\caption{\label{fig:test1-sphere}The upper two pictures are the results of roe scheme without flux limiter. The following two pictures are the results of roe scheme with flux limiter.}
\end{figure}
Without using a flux limiter from 2.53, the results will be volatile. The roe scheme will cause invalid output Nan after 0.5 days of running. It is probably due to a negative height term in the solution obtained by the NN, which leads to a Nan result when calculating the numerical flux. 

\begin{figure}[H]
\centering
\includegraphics[width=1\textwidth]{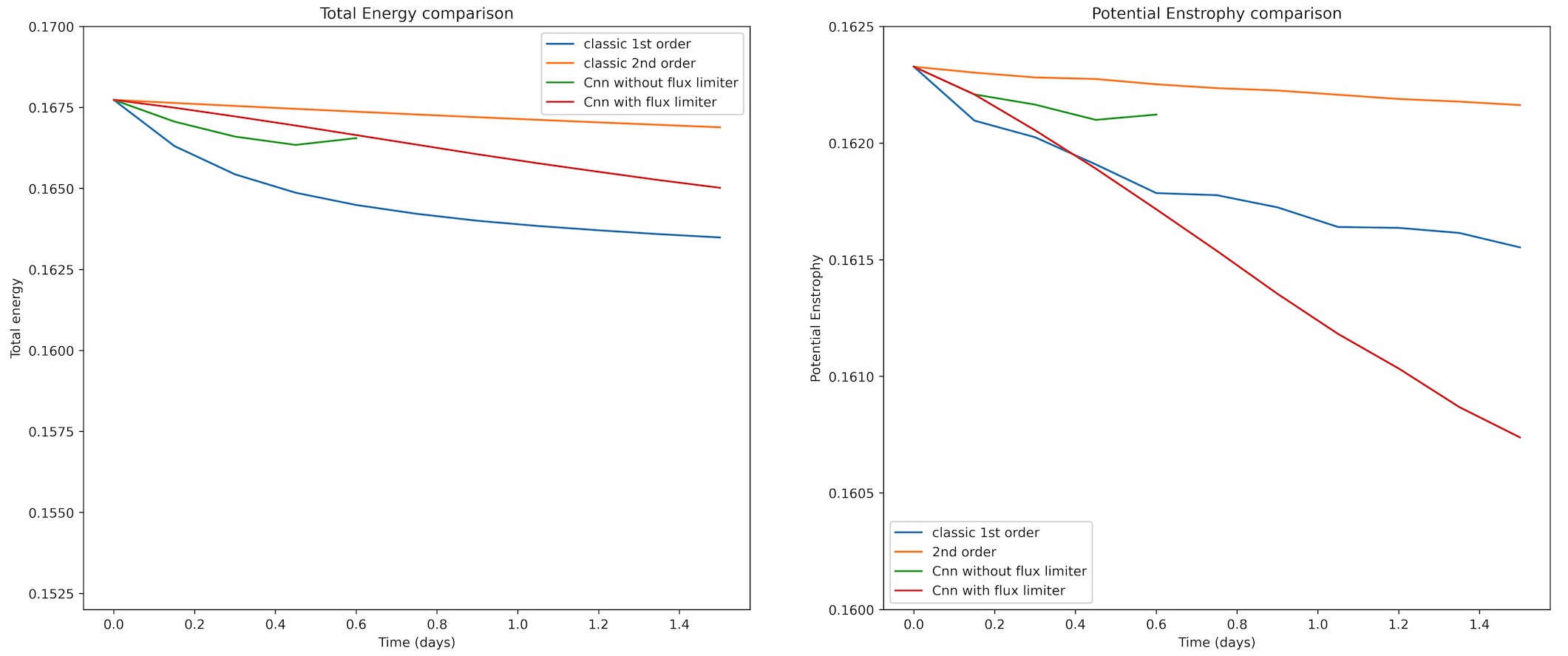}
\caption{\label{fig:Engergy1-sphere}Evolution of the conservation of total energy (left) and potential enstrophy (right) on $100\times50$ mesh grid for Rossby–Haurwitz wave test using second NN approach.}
\end{figure}
After energy testing, a classic first-order solver with $100\times50$ resolution will reduce the total energy from 0.1674 to 0.1638 and second-order to 0.1671 after one day. At the same time, the NN without flux limiter occurs with invalid values from 0.5 days. In addition, the NN with flux limiter performs better in total energy, but the potential entropy loss is severe. It indicates that more potential energy is converted to kinetic energy, and the solver error gets inaccuracy, which is consistent with the results in Figure 25.

We also give an example in the following Figure 27. The "Center Value" represents the cell average used at first-order precision. The "Reconstruct" value is the value we reconstructed to the interfaces using the slop limiter, and "CNN" denotes the boundary values using CNN reconstruction. Furthermore, "True" indicates that we used the continuous initial condition to obtain the actual value of the boundary.
\begin{figure}[H]
\centering
\includegraphics[width=0.89\textwidth]{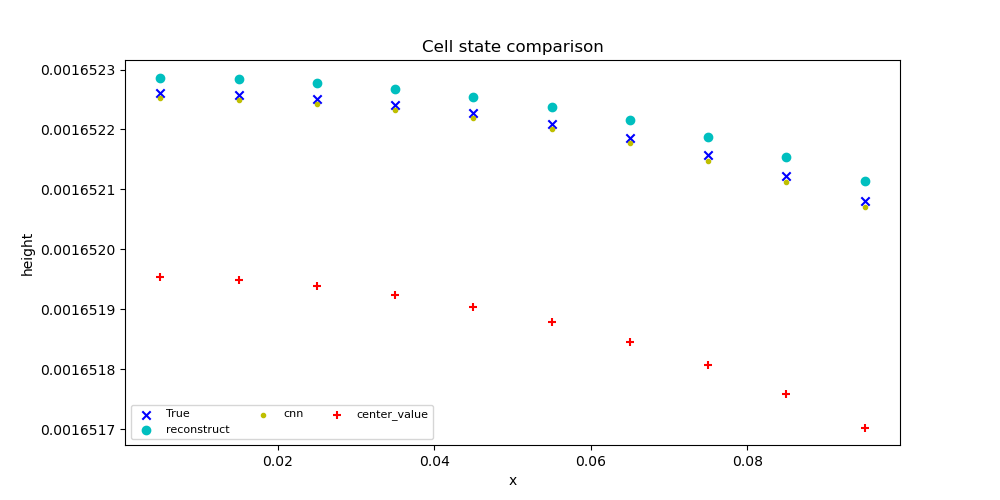}
\caption{\label{fig:cell_state_test}Comparison of the cell state using different reconstruction approaches.}
\end{figure}
We can see that the boundary values obtained using CNN are already very close to the true values and are significantly better than those obtained using the slop limiter.

In general, this NN structure is not feasible. Although the interpolated value at the boundary is close to the actual value, there may be intermittent discontinuities in the interpolated value resulting in an unstable output value. In addition, although the results become continuous after using the flux limiter, the values are severely dissipated, and the results are worse than the first-order scheme.

\subsection{Calculate boundary states directly from CNN}
\paragraph{One dimensional}~\\
For one dimensional case, we use a Gaussian process~\cite{ref28} for the training set as the initial condition to generate a large number of samples with high resolution (2048) and then solve them using the second-order solver in Chapter 3, which are shown in Figure 28. Then we coarsened the obtained numerical solutions to low resolution 64($32\times$), 128($16\times$) and 256($8\times$) after taking their $nT = 2560$ steps.

\begin{figure}[H]
\centering
\includegraphics[width=1.0\textwidth]{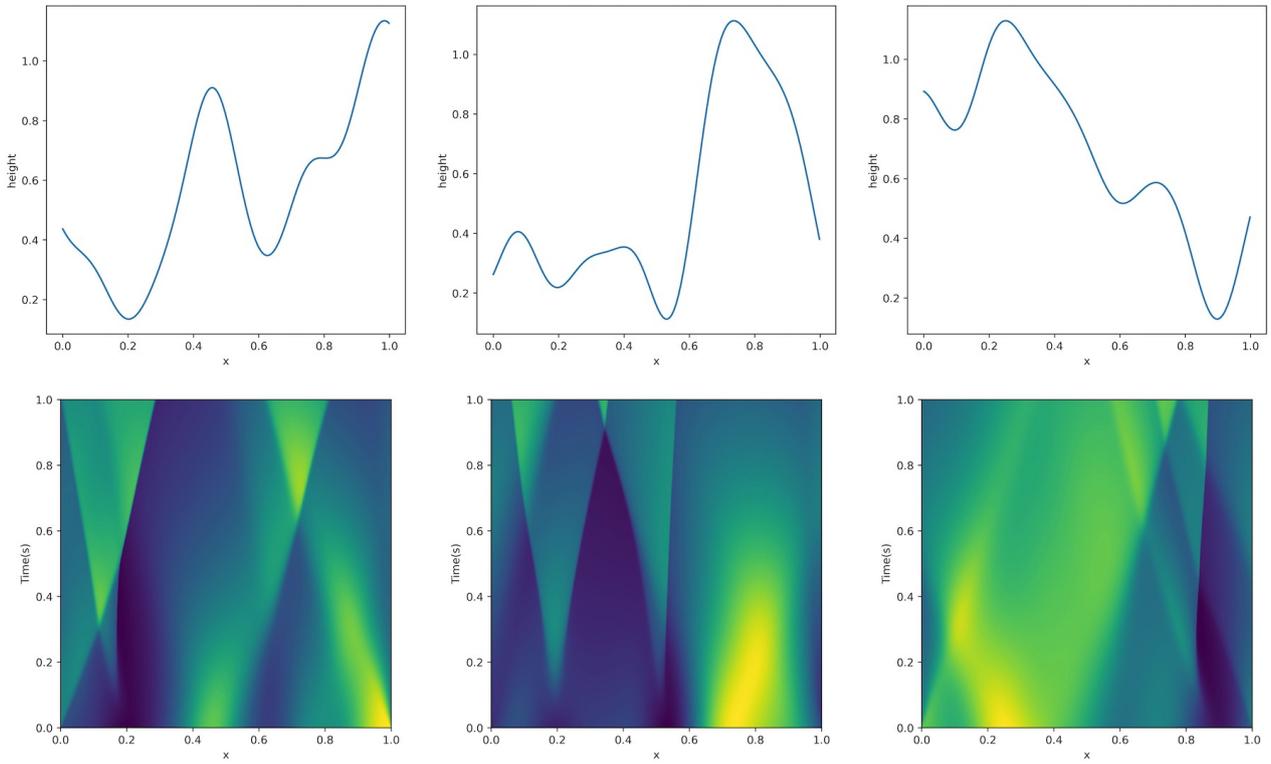}
\caption{\label{fig:swe1d_sample}One dimensional training set generated by 1D gaussian process.}
\end{figure}

After 20 epoch training, we obtained neural networks for Coarsening $8\times$, $16\times$, and $32\times$, respectively. The NN embedded solver's results are shown in Fig. 29.
\begin{figure}[H]
\centering
\includegraphics[width=1.0\textwidth]{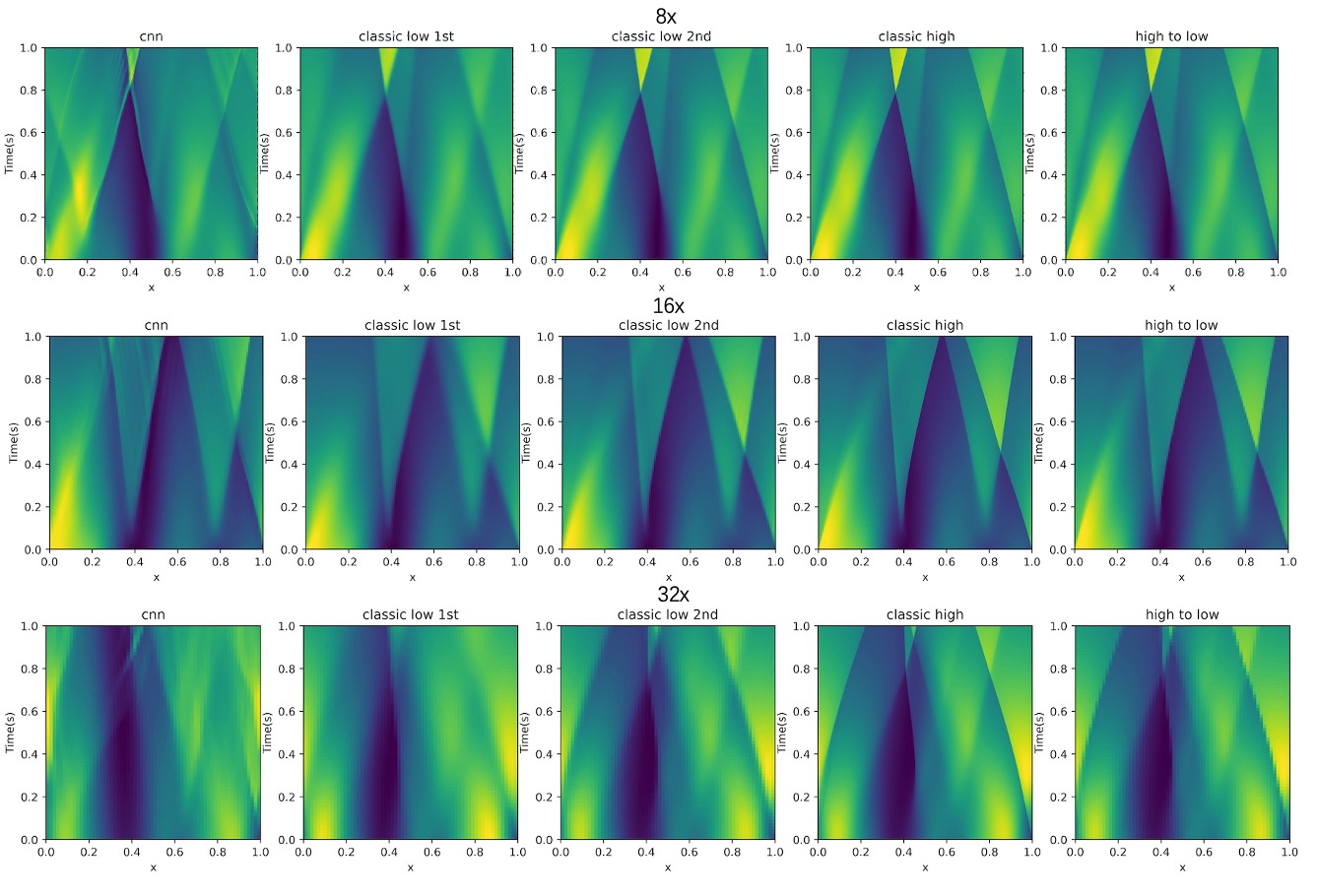}
\caption{\label{fig:test2-1d}Solution of one dimensional ML solver using third NN approach.}
\end{figure}

We can see that the overall result of this NN approach-based solver performs well running for 1 second, but there are still some areas with a bit of noise that makes the result inaccurate. Our target solution after training is the solution of the rightmost high-precision coarsening to a low-precision grid, which shows sharper than the second classic solver.

\begin{figure}[H]
\centering
\includegraphics[width=1\textwidth]{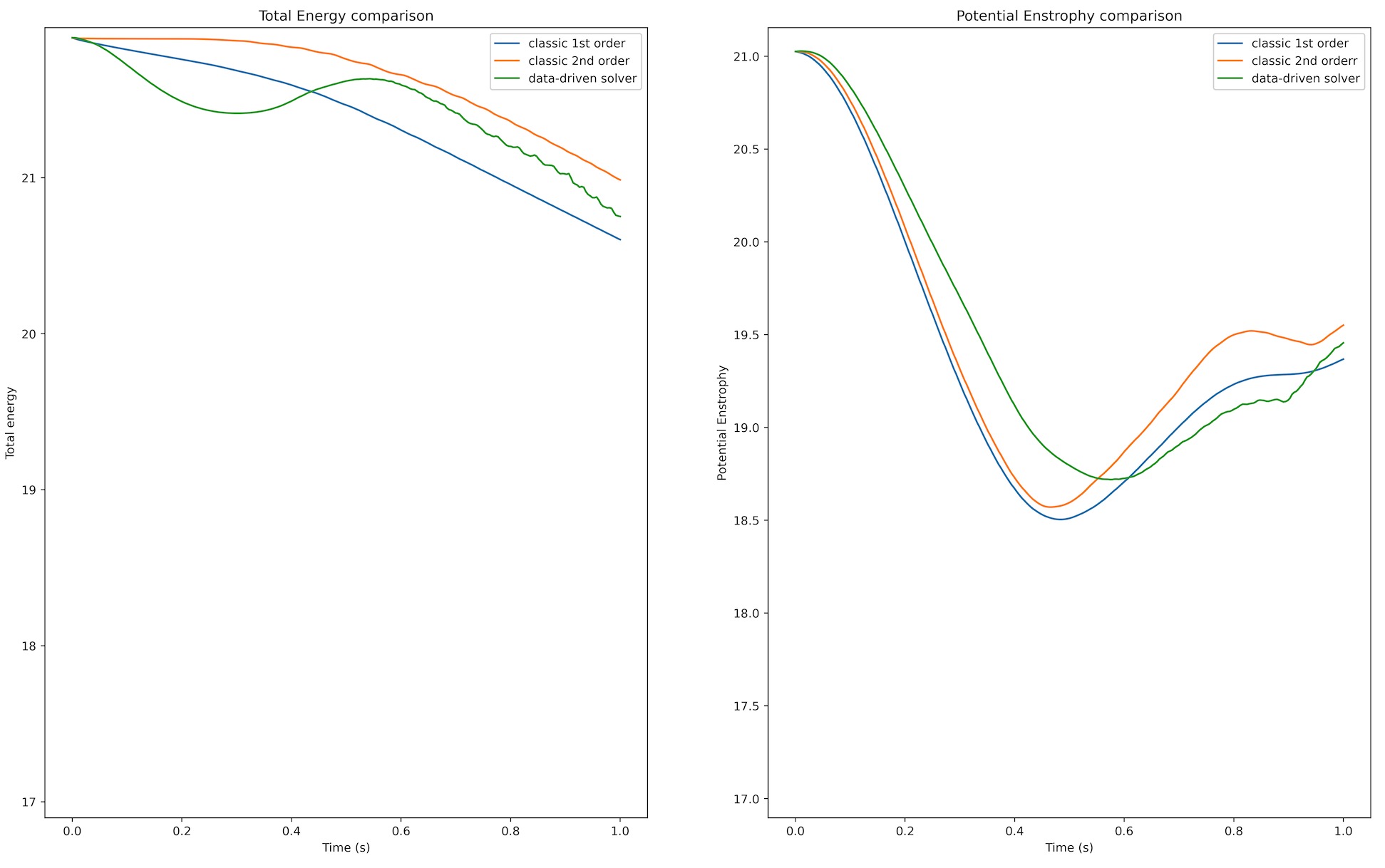}
\caption{\label{fig:Engergy1-sphere}Evolution of the conservation of total energy (left) and potential enstrophy (right) on 128 resolution grid using third NN approach.}
\end{figure}
The above figure shows that using CNN to output the boundary value directly is much more reasonable than the first NN method. The total and potential energy are similar to the classic first-order and second-order schemes. Still, some noise may lead to slightly inaccurate results, but this method is generally feasible.

\paragraph{SWE on planar}~\\
In the two-dimensional case, we use 2D Perlin noise~\cite{ref30} as the initial condition (Figure 31). Moreover, we ensure that the height and momentum components are within a reasonable range. We use NN to output one more momentum term $hv$. Afterward, similar to the one-dimensional case, the two-dimensional solver is run under $1024\times1024$ resolution and then coarsened by a factor of 8 and 16 to a low-resolution mesh grid for 2560 time steps.

\begin{figure}[H]
\centering
\includegraphics[width=1.0\textwidth]{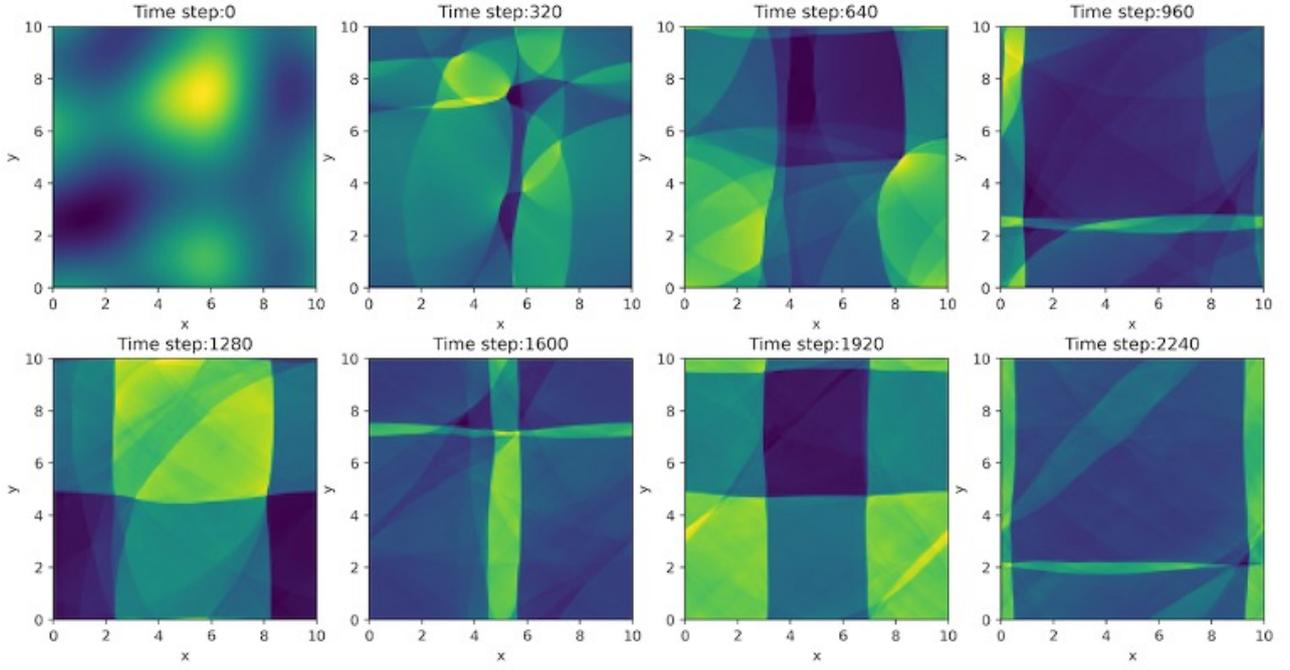}
\caption{\label{fig:trainset2-2d}Two dimensional training set generated by 2D Perlin noise.}
\end{figure}

Note that, for the two-dimensional Neural Network, we use CNN to reconstruct the center value to the four boundaries of each cell directly, i.e., $Q_i^{l}$, $Q_i^{u}$, $Q_i^{r}$, $Q_i^{b}$. Then we use $Q_i^{l}$, $Q_{i-1}^{r}$ and $Q_i^{b}$, $Q_{i-1}^{u}$ to calculate the flux $F_{i}^{l}$ and $F_{i}^{b}$. According to the FVM conservation law we can know $F_{i}^{r}=-F_{i+1}^{r}$ and $F_{i}^{u}=-F_{i+1}^{b}$.

\begin{figure}[H]
\centering
\includegraphics[width=0.9\textwidth]{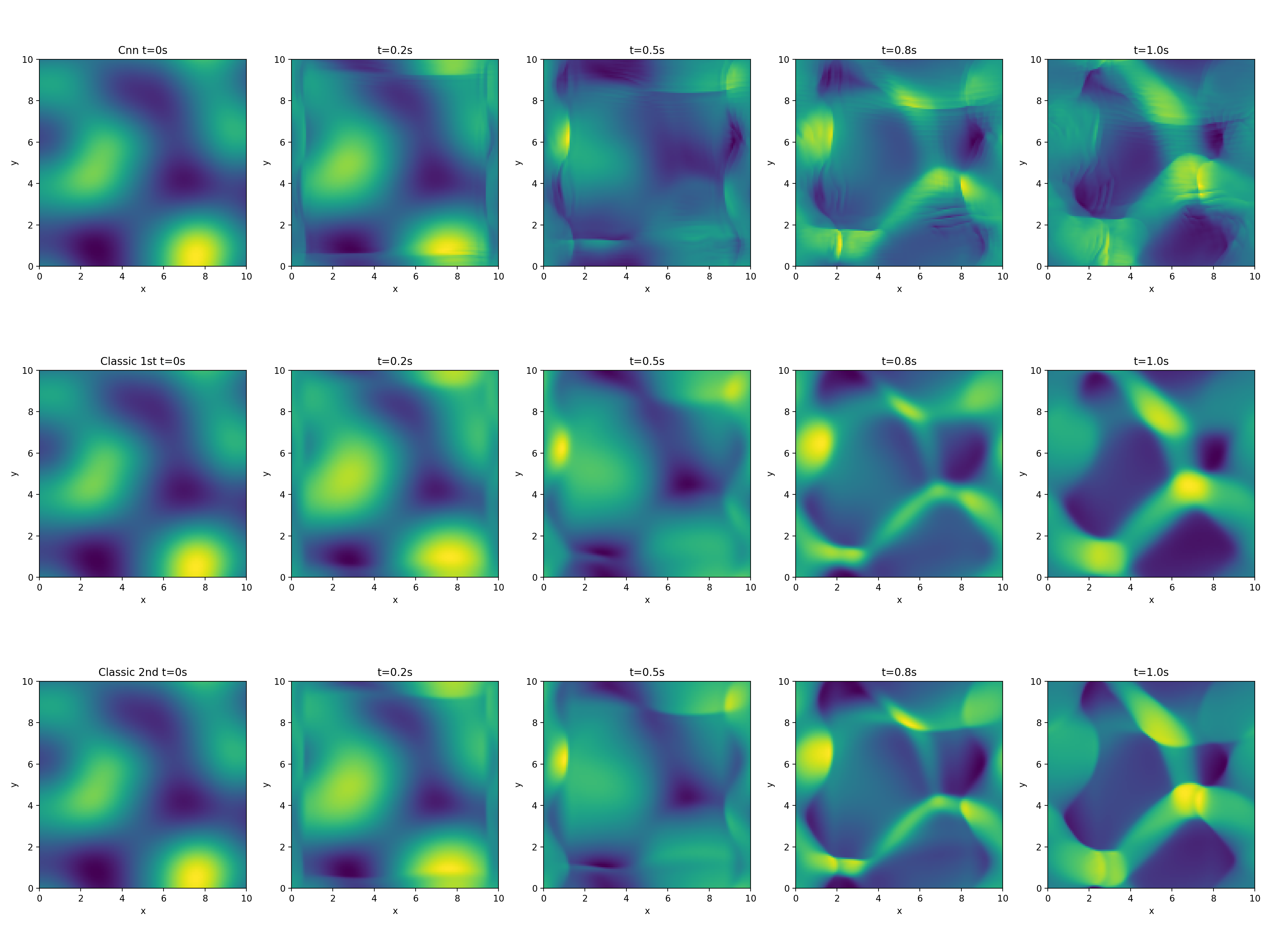}
\caption{\label{fig:test2-2d-8}Solution of two dimensional ML solver using third NN approach($8\times$).}
\end{figure}

\begin{figure}[H]
\centering
\includegraphics[width=0.9\textwidth]{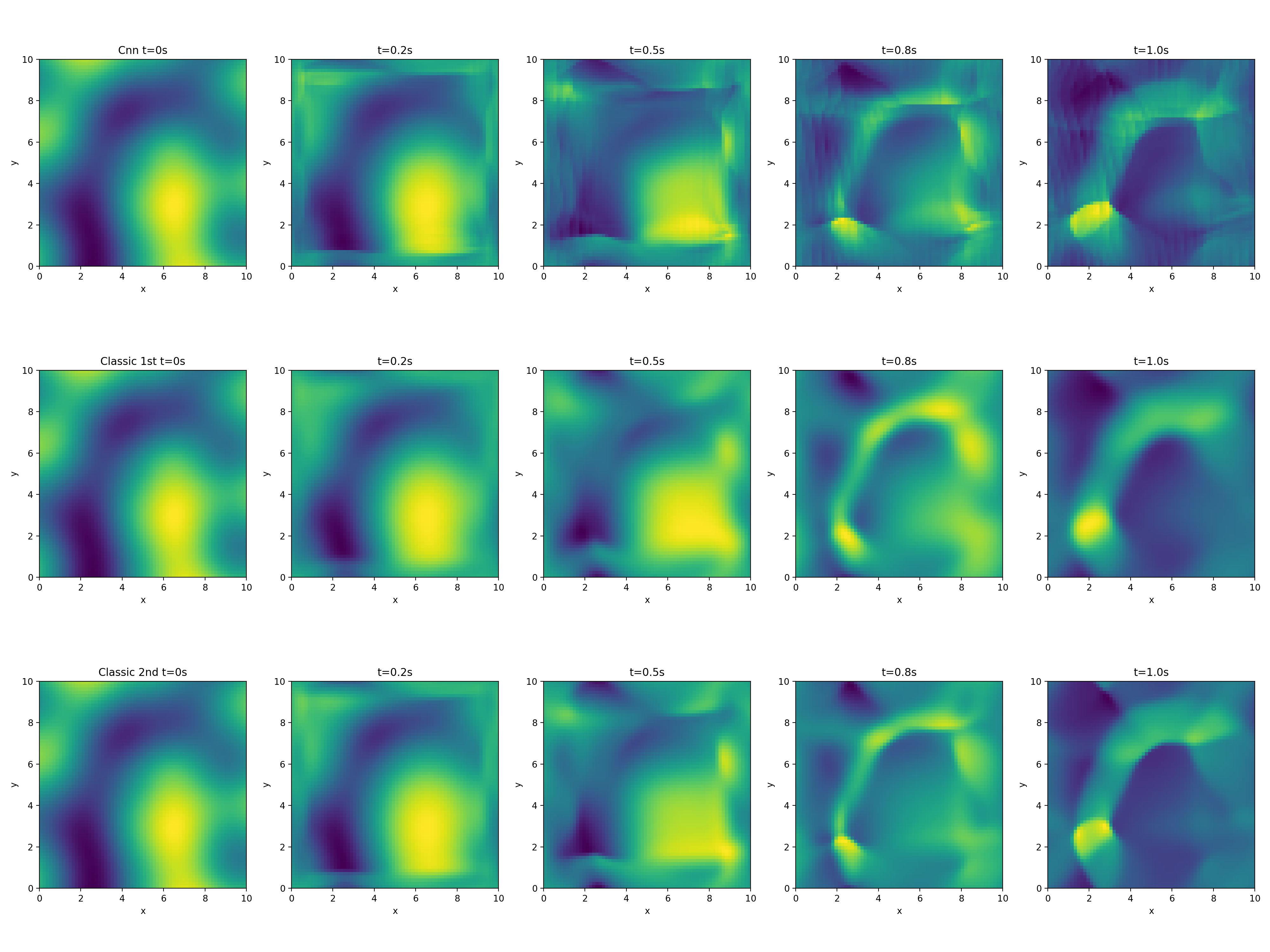}
\caption{\label{fig:test2-2d-16}Solution of two dimensional ML solver using third NN approach($16\times$).}
\end{figure}

After training ten epochs with ten different initial conditions, we find that the results are similar to the one-dimensional case. Our two-dimensional ML solver results are correct in general. Figures 32 and 33 show that the solution after high-resolution coarsening contains more details, which coincides with the solution after CNN interpolation in Figure 31, which cannot be obtained from the low-resolution classic solver.

\begin{figure}[H]
\centering
\includegraphics[width=1\textwidth]{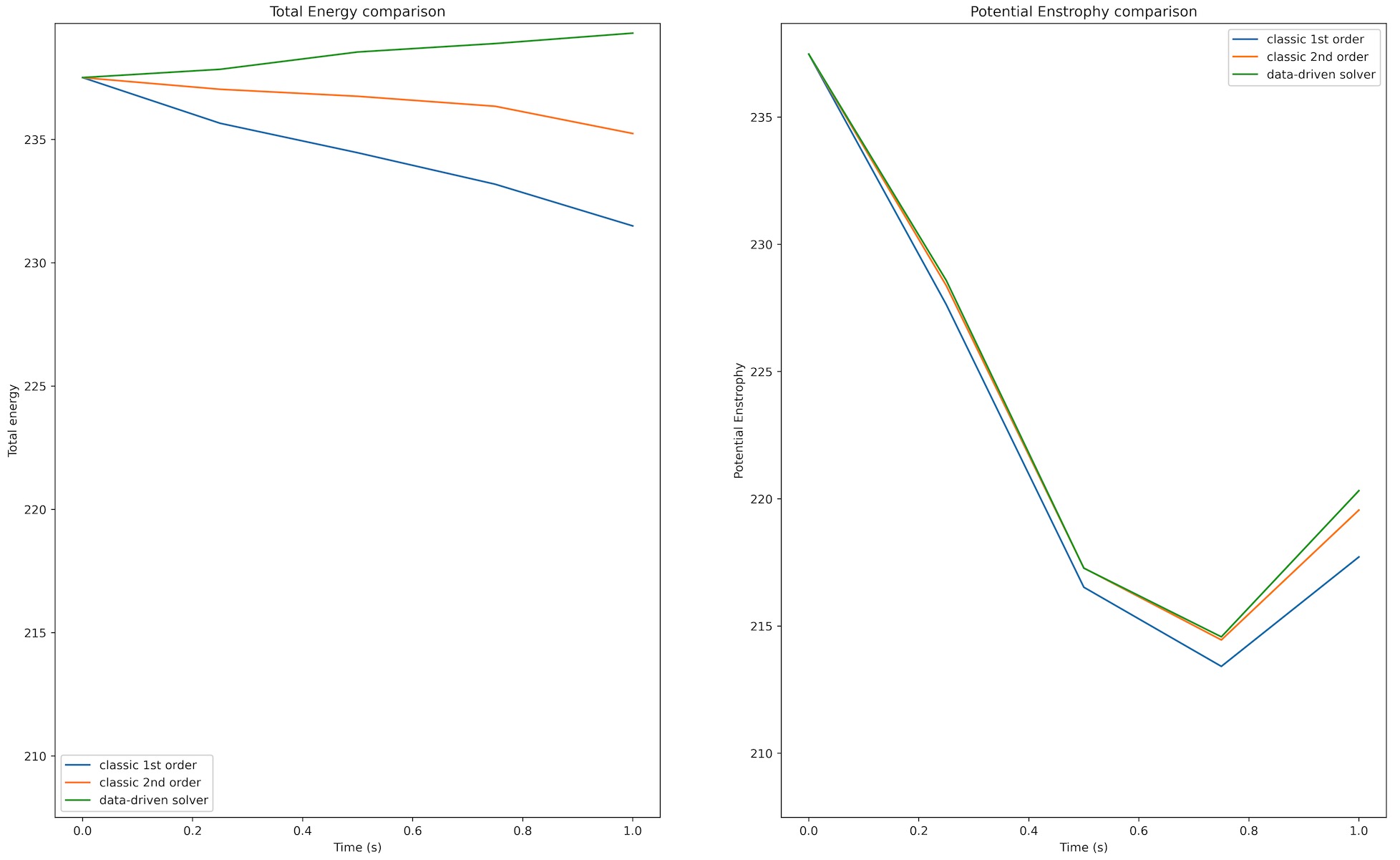}
\caption{\label{fig:Energy2-2d}Evolution of the conservation of total energy (left) and potential enstrophy (right) on $128\times128$ resolution grid using third NN approach.}
\end{figure}
The above figure shows that using CNN to output the boundary value directly is much more reasonable than the first ML method. The total and potential enstrophy changes are close to the classic first-order and second-order schemes. Still, some noise may lead to slightly inaccurate results. After increasing the training epoch, it may be possible to eliminate this noise to obtain more accurate solutions. In general, this NN approach is feasible.

\subsection{Calculate reconstruction slopes with CNN generated linear coefficients}
This section uses the same training sets as in the former.
\paragraph{One dimensional}~\\
After 20 epoch training, we obtained neural networks for Coarsening $8\times$, $16\times$, and $32\times$, respectively. The results are shown in Figure 35,
\begin{figure}[H]
\centering
\includegraphics[width=1.0\textwidth]{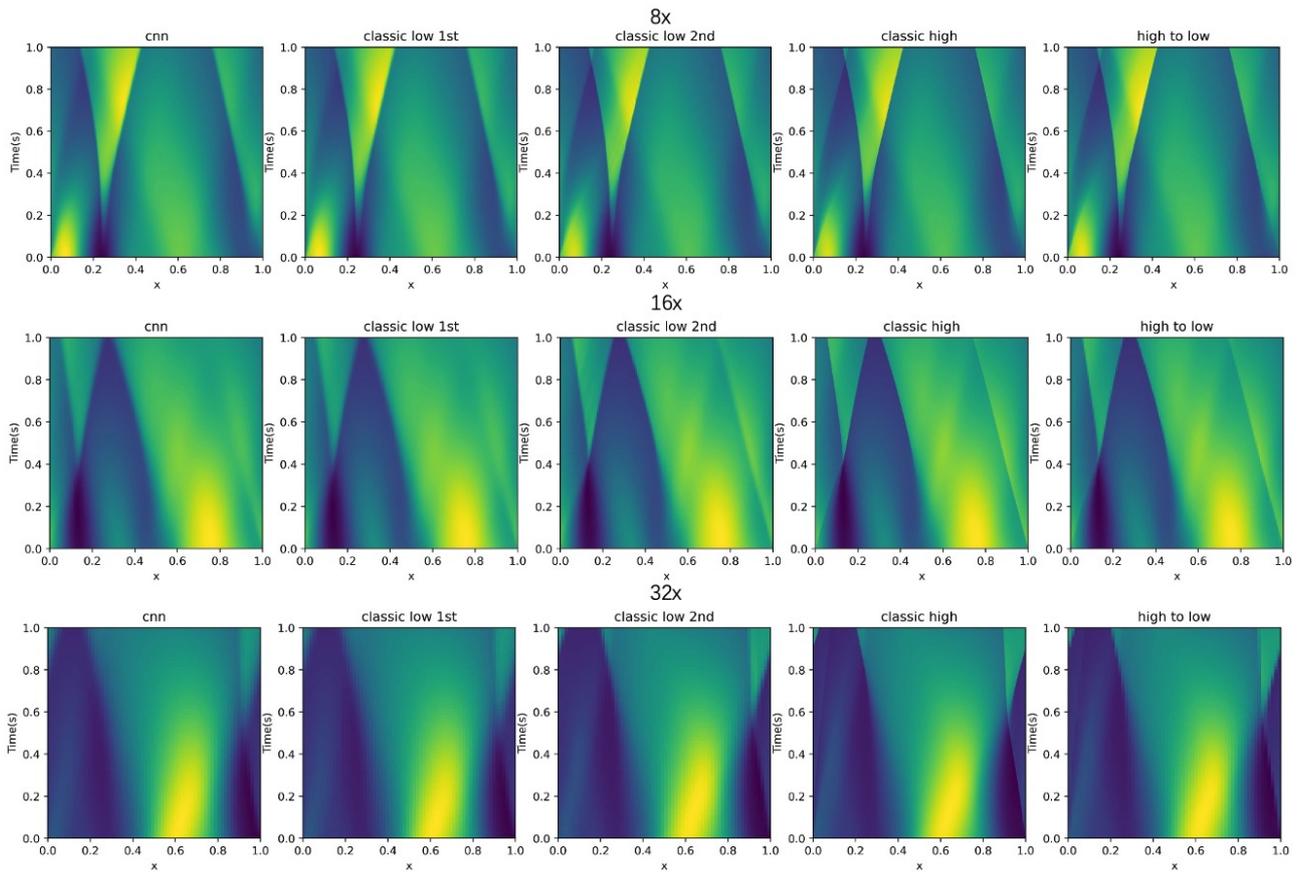}
\caption{\label{fig:test3-1d}Solution of one dimensional ML solver using fourth NN approach.}
\end{figure}
It shows that this NN does not generate noise, and the results are very close to our rightmost target solution.

\begin{figure}[H]
\centering
\includegraphics[width=1\textwidth]{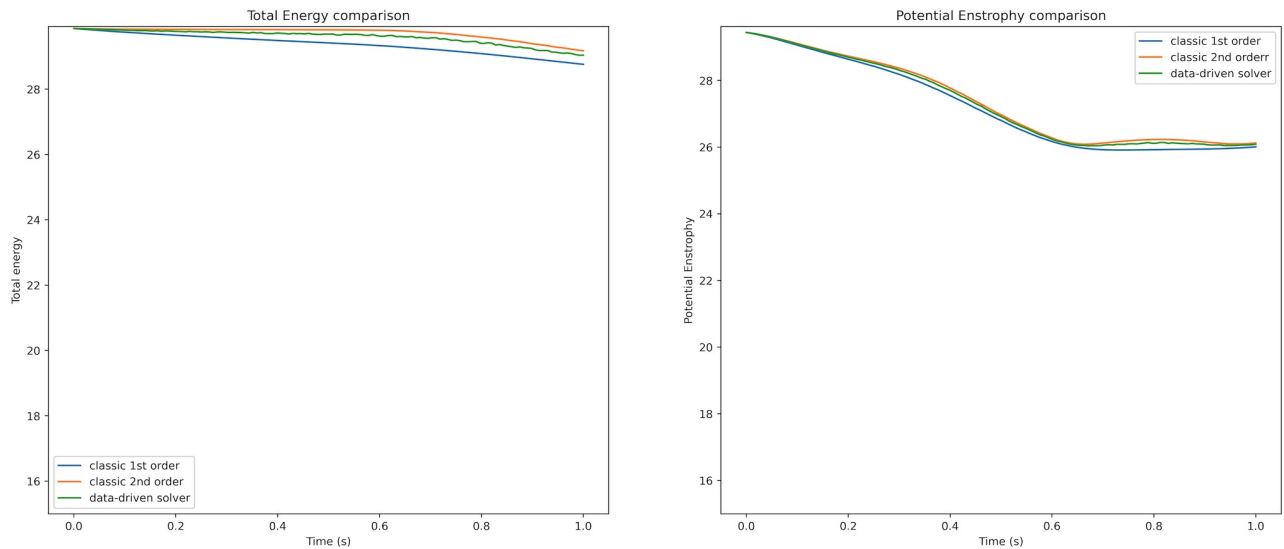}
\caption{\label{fig:Energy3-1d}Evolution of the conservation of total energy (left) and potential enstrophy (right) on 128 resolution grid using fourth NN approach.}
\end{figure}
Figures 35 and 36 show that using CNN to output the reconstruction slop for boundary value is the most stable method. The entropy change curve is similar to the classic solver. What is more, energy loss is better than the first-order classic solver.

\paragraph{SWE on planar}~\\
After training four epochs with ten initial conditions, embedding the neural network into our classical solver running at $128\times128$ resolution for 1s. Compared with the classical solver, we can obtain,

\begin{figure}[H]
\centering
\includegraphics[width=1.0\textwidth]{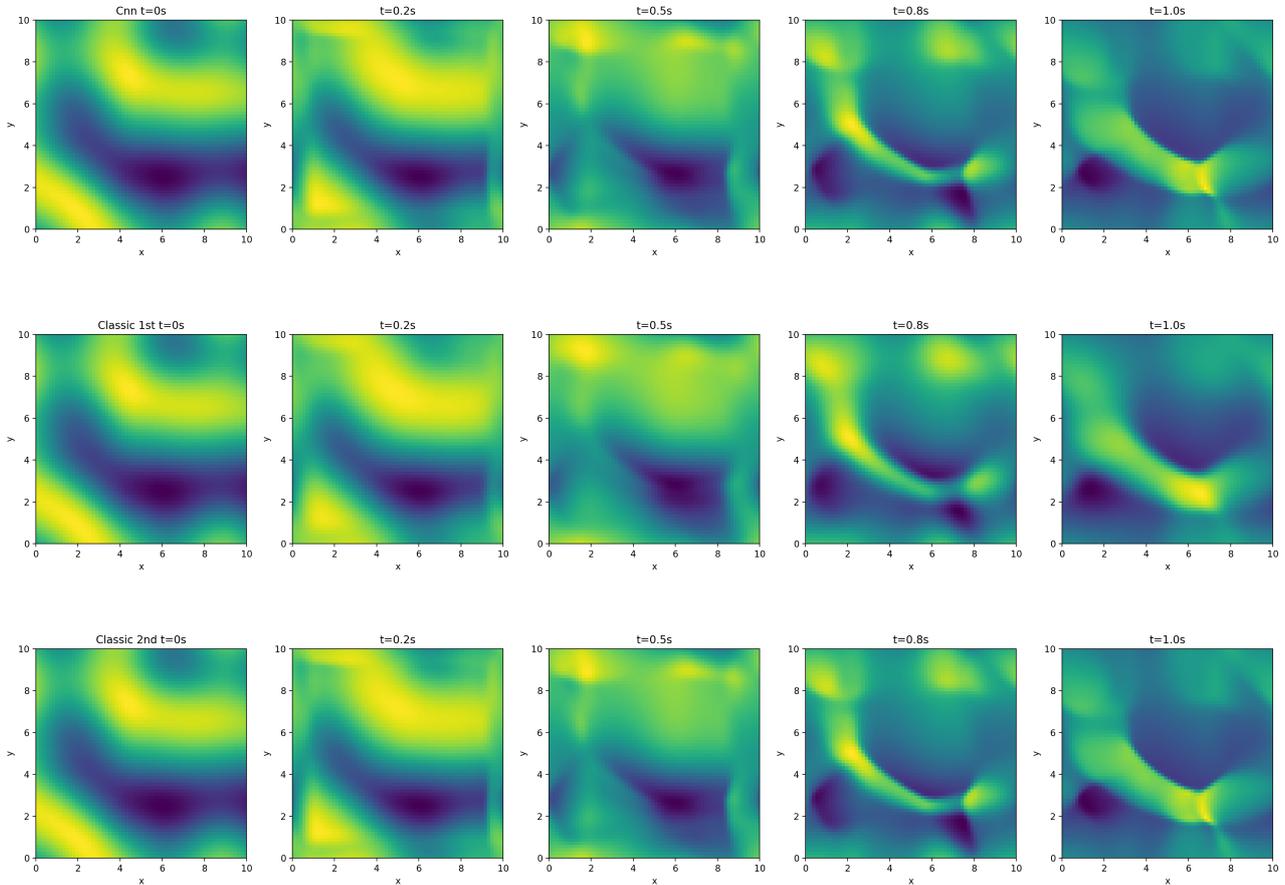}
\caption{\label{fig:test3-2d-16}Solution of two dimensional ML solver using fourth NN approach($16\times$).}
\end{figure}

\begin{figure}[H]
\centering
\includegraphics[width=1\textwidth]{ml/nn3/Energy3-2d.jpg}
\caption{\label{fig:Energy3-2d}Evolution of the conservation of total energy (left) and potential enstrophy (right) on $128\times128$ resolution grid using fourth NN approach.}
\end{figure}
Figure 37 shows that the ML solver results are sharper than the classic first-order solver. Moreover, the potential Enstrophy of the ML-based solver is already very close to that of the second-order solver.

\paragraph{SWE on sphere}~\\
In the sphere case, we use NN to output one more alpha set for $hw$. After training four epochs, the obtained neural network is embedded into the classic solver and run for 3s at $250\times125$ resolution to obtain the following results.
\begin{figure}[H]
\centering
\includegraphics[width=1.0\textwidth]{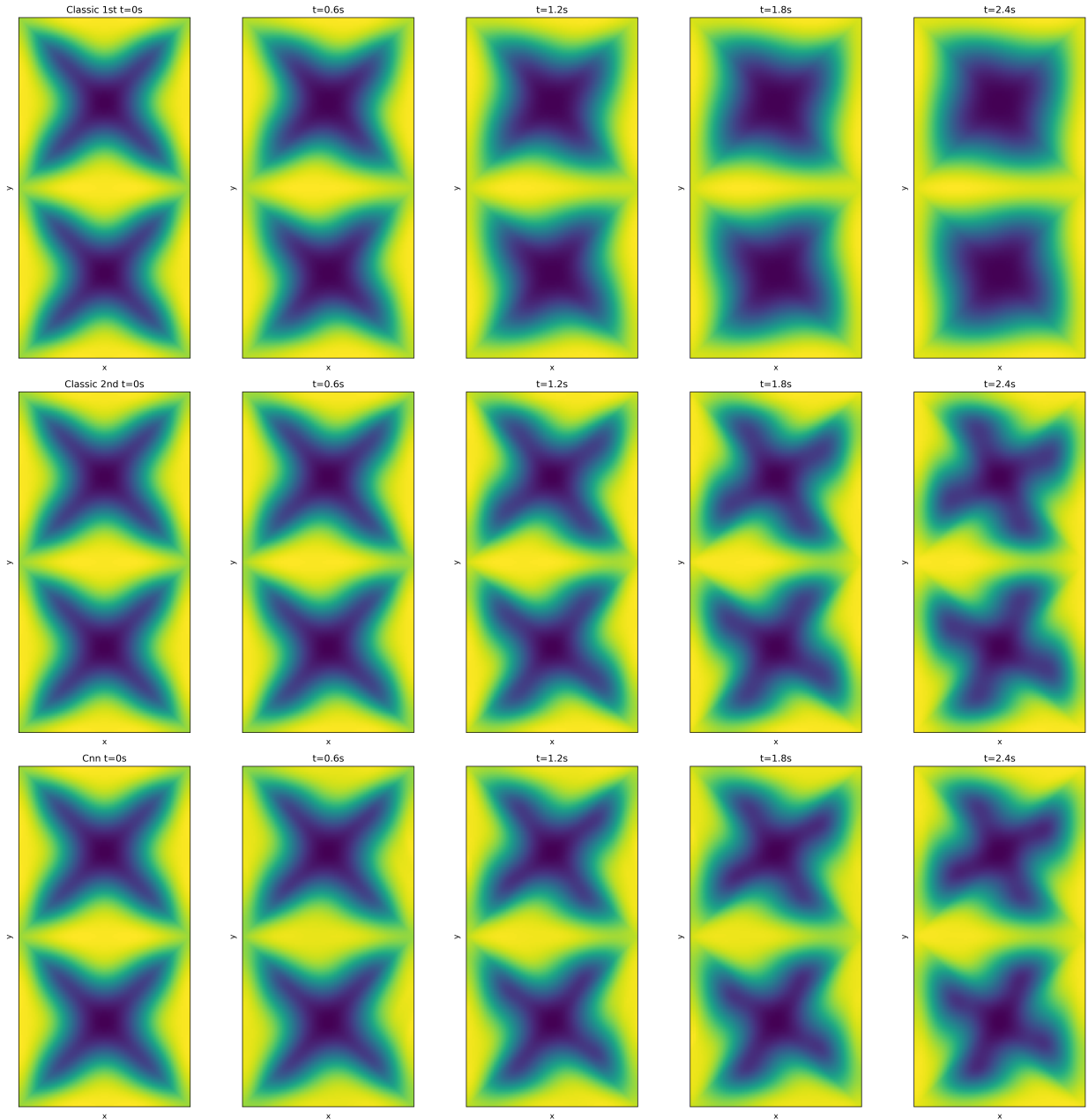}
\caption{\label{fig:test3-sphere-10}Solution of spherical ML solver using fourth NN approach($10\times$) on computational grid.}
\end{figure}

After running for three days, we find that the classic first-order results are blurred due to the lack of precision, but the second-order and ML-based solvers perform well. Moreover, the ML solver has stable performance and no noise.

\begin{figure}[H]
\centering
\includegraphics[width=1\textwidth]{ml/nn3/Energy3-sphere.jpg}
\caption{\label{fig:Energy3-sphere}Evolution of the conservation of total energy (left) and potential enstrophy (right) on $250\times125$ resolution grid using fourth NN approach.}
\end{figure}

The energy loss in Fig. 40 matches the results in Fig. 39. The classic first-order solver has the most severe energy loss, which makes its solution ambiguous. ML solver has a loss between the first and second order, and the results are accurate. This NN approach is the best of the four, with stable and accurate results. If the training time increases and the number of training sets increases, the results may outperform the second-order accuracy solver.

\section{Conclusions and Future Work}
\subsection{Conclusions}
To conclude this thesis, we implement first and second-order one/two-dimensional Euler equations and shallow water equations, in addition to the spherical shallow water equations, which include several numerical fluxes schemes and flux limiters under the torch and dace framework. After testing, our classic solver performs much better than the Pyclaw solver. Then we try to embed the ML into the classic solver. Although much work has been proposed in data-driven-based accelerated CFD solvers, simulations based on shallow water equations still need to be explored. We have developed four data-driven based solvers for the shallow water equation.

Experiments show that the first method, which uses CNN to output the flux directly, is not feasible. Furthermore, the second NN approach, using the Learned coefficients method to calculate the boundary values, does not perform well due to the intermittent discontinuity of the boundary values obtained by interpolation, resulting in unstable results. In the third method, we use CNN to output the boundary values directly. There may be some noise due to insufficient training epoch or limited train set, but this method is feasible overall. In the fourth method, we compute reconstruction slopes with CNN to generate linear coefficients, which have the best performance and can guarantee at least first-order accuracy.

\subsection{Future Work}
There is still space for further improvements, such as in inputs and model architectures. In the future, the first thing is to increase the training set with the training period in the second method. Our method is based on the WENO method for nonlinear interpolation to obtain the boundary values, which can theoretically achieve arbitrary accuracy results

In addition, we also need to find an optimal number of stencils, which theoretically can improve the accuracy. In the previous experiments, we used three stencils. We can also use the Learned Correction method~\cite{ref9}, closer in spirit to LES modeling, to implement the ml solver, which uses CNN to directly output the flux or cell state. We have tried this method but have yet to solve the stability problem.

Furthermore, we can continue to embed the data-driven method into Euler's equation. The Euler equation requires that the density and energy terms be positive, so more constraints should be considered.

\section*{Data Availability}
Source code for our models, including learned components, and train set generator are available at GitHub (\url{https://github.com/baaiy0610/master-thesis}).

\newpage
\thispagestyle{empty}
\listoffigures
\listoftables
\newpage


\begin{thebibliography}{99} 
\bibitem{ref1}Kämpf Jochen (2009) Ocean modelling for beginners: Using open-source software. Berlin: Springer. 

\bibitem{ref2}LeVeque, R.J. (2011) Finite volume methods for hyperbolic problems. Cambridge: Cambridge Univ. Press. 

\bibitem{ref3}Mishra, S. (2018) “A machine learning framework for data driven acceleration of computations of Di Erential equations,” Mathematics in Engineering, 1(1), pp. 118–146. Available at: https://doi.org/10.3934/mine.2018.1.118. 

\bibitem{ref4}Bar-Sinai, Y. et al. (2019) “Learning data-driven discretizations for partial differential equations,” Proceedings of the National Academy of Sciences, 116(31), pp. 15344–15349. Available at: https://doi.org/10.1073/pnas.1814058116. 

\bibitem{ref5}Magiera, J. et al. (2020) “Constraint-aware neural networks for Riemann problems,” Journal of Computational Physics, 409, p. 109345. Available at: https://doi.org/10.1016/j.jcp.2020.109345. 

\bibitem{ref6}Maddu, Suryanarayana, et al. "STENCIL-NET: Data-driven solution-adaptive discretization of partial differential equations." arXiv preprint arXiv:2101.06182 (2021).

\bibitem{ref7}Shu, C.-W. (1998) “Essentially non-oscillatory and weighted essentially non-oscillatory schemes for hyperbolic conservation laws,” Lecture Notes in Mathematics, pp. 325–432. Available at: https://doi.org/10.1007/bfb0096355. 

\bibitem{ref8}Weyn, J.A., Durran, D.R. and Caruana, R. (2020) “Improving data-driven global weather prediction using deep convolutional neural networks on a cubed sphere.” Available at: https://doi.org/10.1002/essoar.10502543.1. 

\bibitem{ref9}Kochkov, D. et al. (2021) “Machine learning–accelerated computational fluid dynamics,” Proceedings of the National Academy of Sciences, 118(21). Available at: https://doi.org/10.1073/pnas.2101784118. 

\bibitem{ref10}Vreugdenhil, C.B. (2011) Numerical methods for shallow-water flow. Dordrecht: Springer. 

\bibitem{ref11}Tec-Science (2021) Derivation of the Euler equation of motion (conservation of momentum) - tec-science, tec. Available at: http://www.tec-science.com/mechanics/gases-and-liquids/derivation-of-the-euler-equation-of-motion-conservation-of-momentum/ (Accessed: November 22, 2022). 

\bibitem{ref12}Courant, R., Friedrichs, K. and Lewy, H. (1986) “Über die Partiellen Differenzengleichungen der mathematischen Physik,” Kurt Otto Friedrichs, pp. 53–95, 10.1007/978-1-4612-5385-3\_7. 

\bibitem{ref13}Press, WH; Teukolsky, SA; Vetterling, WT; Flannery, BP (2007), "Section 10.1.2. Lax Method", Numerical Recipes: The Art of Scientific Computing (3rd ed.), New York: Cambridge University Press, ISBN 978-0-521-88068-8

\bibitem{ref14}Abdoul Kader, M.Y., Rab\'e, B. and Bisso, S. (2022) “A new variant of Rusanov scheme: $\beta-$Rusanov for numerical resolution of shallow water flows,” International Journal of Apllied Mathematics, 35(4). Available at: https://doi.org/10.12732/ijam.v35i4.7. 

\bibitem{ref15}Ambrosi, D. (1995) “Approximation of shallow water equations by Roe's Riemann solver,” International Journal for Numerical Methods in Fluids, 20(2), pp. 157–168. Available at: https://doi.org/10.1002/fld.1650200205. 

\bibitem{ref16}Toro, E.F. (1997). The HLL and HLLC Riemann Solvers. In: Riemann Solvers and Numerical Methods for Fluid Dynamics. Springer, Berlin, Heidelberg. Available at: https://doi.org/10.1007/978-3-662-03490-3\_10

\bibitem{ref17}I. Delis, A. (2002) “Improved application of the HLLE Riemann solver for the shallow water equations with source terms,” Communications in Numerical Methods in Engineering, 19(1), pp. 59–83. Available at: https://doi.org/10.1002/cnm.570. 
%
\bibitem{ref18}Kim, D.-H. and Cho, Y.-S. (2004) “Analysis of shallow-water equations with HLLC approximate Riemann solver,” Journal of Korea Water Resources Association, 37(10), pp. 845–855. Available at: https://doi.org/10.3741/jkwra.2004.37.10.845. 
%
\bibitem{ref19}Balsara, D.S. (2012) “A two-dimensional HLLC Riemann solver for conservation laws: Application to euler and magnetohydrodynamic flows,” Journal of Computational Physics, 231(22), pp. 7476–7503. Available at: https://doi.org/10.1016/j.jcp.2011.12.025. 

\bibitem{ref20}Hu, L. et al. (2022) “Effects of different slope limiters on stratified shear flow simulation in a non-hydrostatic model,” Journal of Marine Science and Engineering, 10(4), p. 489. Available at: https://doi.org/10.3390/jmse10040489. 

\bibitem{ref21}Heun's method (2021) Wikipedia. Wikimedia Foundation. Available at: https://en.wikipedia.org/wiki/Heun\%27s\_method (Accessed: November 23, 2022). 

\bibitem{ref22}Gottlieb, S. and Shu, C.-W. (1998) “Total variation diminishing Runge-Kutta Schemes,” Mathematics of Computation, 67(221), pp. 73–85. Available at: https://doi.org/10.1090/s0025-5718-98-00913-2. 

\bibitem{ref23}Bale, D.S. and LeVeque, R.J. (2001) “Wave propagation algorithms for hyperbolic systems on curved manifolds,” Hyperbolic Problems: Theory, Numerics, Applications, pp. 129–138. Available at: https://doi.org/10.1007/978-3-0348-8370-2\_14. 

\bibitem{ref24}Calhoun, D.A., Helzel, C. and LeVeque, R.J. (2008) “Logically rectangular grids and finite volume methods for pdes in circular and spherical domains,” SIAM Review, 50(4), pp. 723–752. Available at: https://doi.org/10.1137/060664094. 
%
\bibitem{ref25}Williamson, D.L. et al. (1991) “A standard test set for numerical approximations to the shallow water equations in spherical geometry.” Available at: https://doi.org/10.2172/5232139. 

\bibitem{ref26}Ben-Nun, T. et al. (2022) Productive Performance Engineering for weather and climate modeling with python, arXiv.org. Available at: https://arxiv.org/abs/2205.04148v1 (Accessed: November 23, 2022). 

\bibitem{ref27}Liska, R. and Wendroff, B. (2003) “Comparison of several difference schemes on 1D and 2D test problems for the Euler equations,” SIAM Journal on Scientific Computing, 25(3), pp. 995–1017. Available at: https://doi.org/10.1137/s1064827502402120. 

\bibitem{ref28}“Associated Gaussian Processes” (2006) Markov Processes, Gaussian Processes, and Local Times, pp. 551–579. Available at: https://doi.org/10.1017/cbo9780511617997.013. 

\bibitem{ref29}Night (2021) How to use Perlin Noise on a sphere?, DevForum. Available at: https://devforum.roblox.com/t/how-to-use-perlin-noise-on-a-sphere/1545154 (Accessed: November 23, 2022). 
%
\bibitem{ref30}Perlin Noise (2022) Wikipedia. Wikimedia Foundation. Available at: https://en.wikipedia.org/wiki/Perlin\_noise (Accessed: November 23, 2022). 

\bibitem{ref31}Interpolation (scipy.interpolate)\# (no date) Interpolation (scipy.interpolate) - SciPy v1.9.3 Manual. Available at: https://docs.scipy.org/doc/scipy/reference/interpolate.html (Accessed: November 23, 2022). 

\bibitem{ref32}Universal Regridder for geospatial data (no date) xESMF. Available at: https://xesmf.readthedocs.io/en/latest/ (Accessed: November 23, 2022). 

\bibitem{ref33}Chen, C. and Xiao, F. (2008) “Shallow water model on cubed-sphere by multi-moment finite volume method,” Journal of Computational Physics, 227(10), pp. 5019–5044. Available at: https://doi.org/10.1016/j.jcp.2008.01.033. 

\end{thebibliography}
\end{document}